\documentclass[10pt,twocolumn,letterpaper]{article}

\usepackage{cvpr}              % To produce the CAMERA-READY version
\usepackage{hyperref}
\usepackage{adjustbox}
\usepackage{algorithm}
\usepackage{algorithmic}
\usepackage{multirow}
\usepackage{booktabs}
\usepackage{xcolor}
\usepackage{colortbl}
\usepackage{adjustbox}
\definecolor{lightgray}{gray}{0.95}
\definecolor{lightblue}{HTML}{E6F0FF}
\definecolor{lightgreen}{HTML}{E8F5E9}
\definecolor{lightorange}{HTML}{FFF3E0}
\definecolor{bestblue}{HTML}{1565C0}
\definecolor{secondgray}{HTML}{5D5D5D}
\definecolor{headergray}{gray}{0.92}
\definecolor{sectionblue}{HTML}{E3F2FD}
\definecolor{rowgray}{gray}{0.97}
\definecolor{highlightgreen}{HTML}{E8F5E9}
\definecolor{bestblue}{HTML}{1565C0}

\definecolor{cvprblue}{rgb}{0.21,0.49,0.74}
\def\paperID{} % *** Enter the Paper ID here
\def\confName{CVPR}
\def\confYear{2026}

\title{Test-Time Perturbation Learning with Delayed Feedback for Vision-Language-Action Models}
\author{
Zehua Zang$^{1,2}$ 
Xi Wang$^{1,2}$ 
Fuchun Sun$^{3}$ 
Xiao Xu$^{5}$ 
Lixiang Liu$^{1}$ 
Jiahuan Zhou$^{4}$\thanks{Corresponding authors.}~
Jiangmeng Li$^{1}$\footnotemark[1]\\
$^{1}$Institute of Software, Chinese Academy of Sciences, Beijing, China\\
$^{2}$University of Chinese Academy of Sciences, Beijing, China \quad
$^{3}$Tsinghua University, Beijing, China\\
$^{4}$Wangxuan Institute of Computer Technology, Peking University, Beijing, China\\
$^{5}$National Defense University, Beijing, China\\
{\tt\small \{zehua2020,wangxi2024,lixiang,jiangmeng2019\}@iscas.ac.cn,}\\
{\tt\small  fcsun@tsinghua.edu.cn, xuxiao0825@gmail.com, jiahuanzhou@pku.edu.cn}
}

\begin{document}
\maketitle

% \footnotetext{* Corresponding authors.}
\begin{abstract}
Vision-Language-Action models (VLAs) achieve remarkable performance in sequential decision-making but remain fragile to subtle environmental shifts, such as small changes in object pose. We attribute this brittleness to trajectory overfitting, where VLAs over-attend to the spurious correlation between actions and entities, then reproduce memorized action patterns. We propose Perturbation learning with Delayed Feedback (PDF), a verifier-free test-time adaptation framework that improves decision performance without fine-tuning the base model. PDF mitigates the spurious correlation through uncertainty-based data augmentation and action voting, while an adaptive scheduler allocates augmentation budgets to balance performance and efficiency. To further improve stability, PDF learns a lightweight perturbation module that retrospectively adjusts action logits guided by delayed feedback, correcting overconfidence issue. Experiments on LIBERO (+7.4\% success rate) and Atari (+10.3 human normalized score) demonstrate consistent gains of PDF in task success over vanilla VLA and VLA with test-time adaptation, establishing a practical path toward reliable test-time adaptation in multimodal decision-making agents. The code is available at \href{https://github.com/zhoujiahuan1991/CVPR2026-PDF}{https://github.com/zhoujiahuan1991/CVPR2026-PDF}.
\end{abstract}

\section{Introduction}\label{sec:intro}

%[AS] Vision-Language-Action models (VLAs)~\cite{spatialvla,jarvisvla,CoT-VLA,3D-VLA,FeedTTA} have emerged as a powerful paradigm for sequential decision-making, showing strong performance in robotic manipulation~\cite{openvla,pi0,openvla-oft} and visually grounded control tasks~\cite{gato,jat}. By jointly modeling visual perception, language understanding, and action generation, VLAs translate high-level instructions into executable actions through large-scale multimodal pretraining, marking a promising step toward general-purpose embodied intelligence.

Vision-Language-Action models (VLAs)~\cite{spatialvla,jarvisvla,CoT-VLA,3D-VLA,FeedTTA} have emerged as a powerful paradigm for sequential decision-making, demonstrating strong performance in robotic manipulation~\cite{openvla,pi0,openvla-oft} and visually grounded control tasks~\cite{gato,jat}. By jointly modeling visual, language, and action, VLAs translate instructions into executable actions through multimodal pretraining, marking a promising step toward general-purpose embodied intelligence.

\begin{figure}[t]
    \centering
    \includegraphics[width=1.0\linewidth]{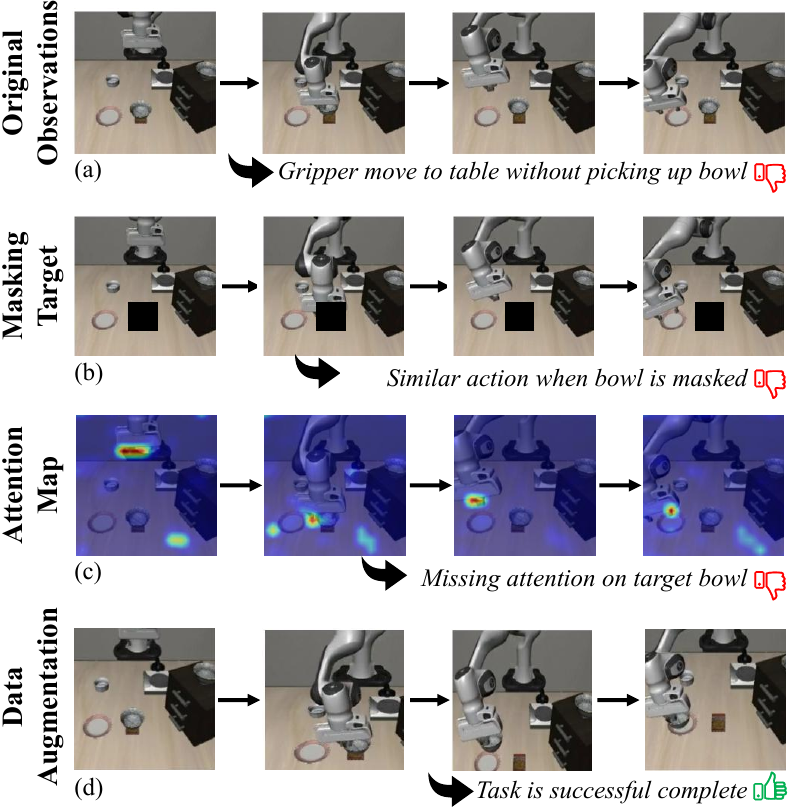}
    \caption{
    %[AS] Evidence of trajectory overfitting and the effectiveness of data augmentation. In the first row, the robot gripper closely imitates the expert trajectories, regardless of whether it successfully grasps the target object. In the second row, even when the target object is masked, the gripper continues to reproduce actions similar to the expert’s trajectories. The attention maps in the third row reveal that the gripper overlooks the target during decision-making. In contrast, the final row shows that after applying data augmentations to the pixel observations, the gripper successfully refocuses on the target and makes correct decisions.
    Evidence of trajectory overfitting and the effectiveness of data augmentation. In the first row, the gripper imitates expert trajectories regardless of task success. In the second row, it still reproduces similar actions when the target is masked. The third row shows that the gripper overlooks the target in attention maps. In contrast, after data augmentation, the gripper refocuses on the target and makes correct decisions.
    }
    \label{fig:motivation}
\end{figure}

Despite these impressive capabilities, VLAs remain fragile under even minor, semantically insignificant environmental shifts, such as slight changes in object pose, often leading to severe performance degradation or task failure~\cite{LIBERO-Plus}. This brittleness reveals a gap in test-time decision performance. To further elucidate this issue, we conduct diagnostic experiments that expose an intriguing behavioral bias. As shown in Figure~\ref{fig:motivation}(a), when instructed to grasp a bowl, the robotic arm may fail to complete the action but continues moving toward the plate, suggesting an overreliance on learned spatial or visual correlations rather than true semantic understanding. Even when the target bowl is masked (Figure~\ref{fig:motivation}(b)), the agent exhibits similar motions, indicating that the policy is not grounded in the visual semantics of the goal. We refer to this failure mode as trajectory overfitting, where VLAs memorize trajectory-specific visual or contextual patterns—such as gripper appearance or background textures—associated with success during training (Figure~\ref{fig:motivation}(c)), and reproduce suboptimal action traces when inputs resemble past trajectories. As shown in Figure~\ref{fig:motivation}(d), applying controlled data augmentation to pixel observations can restore correct execution, suggesting that perturbing the input distribution helps mitigate trajectory overfitting. This motivates improving decision performance via perturbation learning without costly fine-tuning.

\begin{figure}
    \centering
    \includegraphics[width=1.0\linewidth]{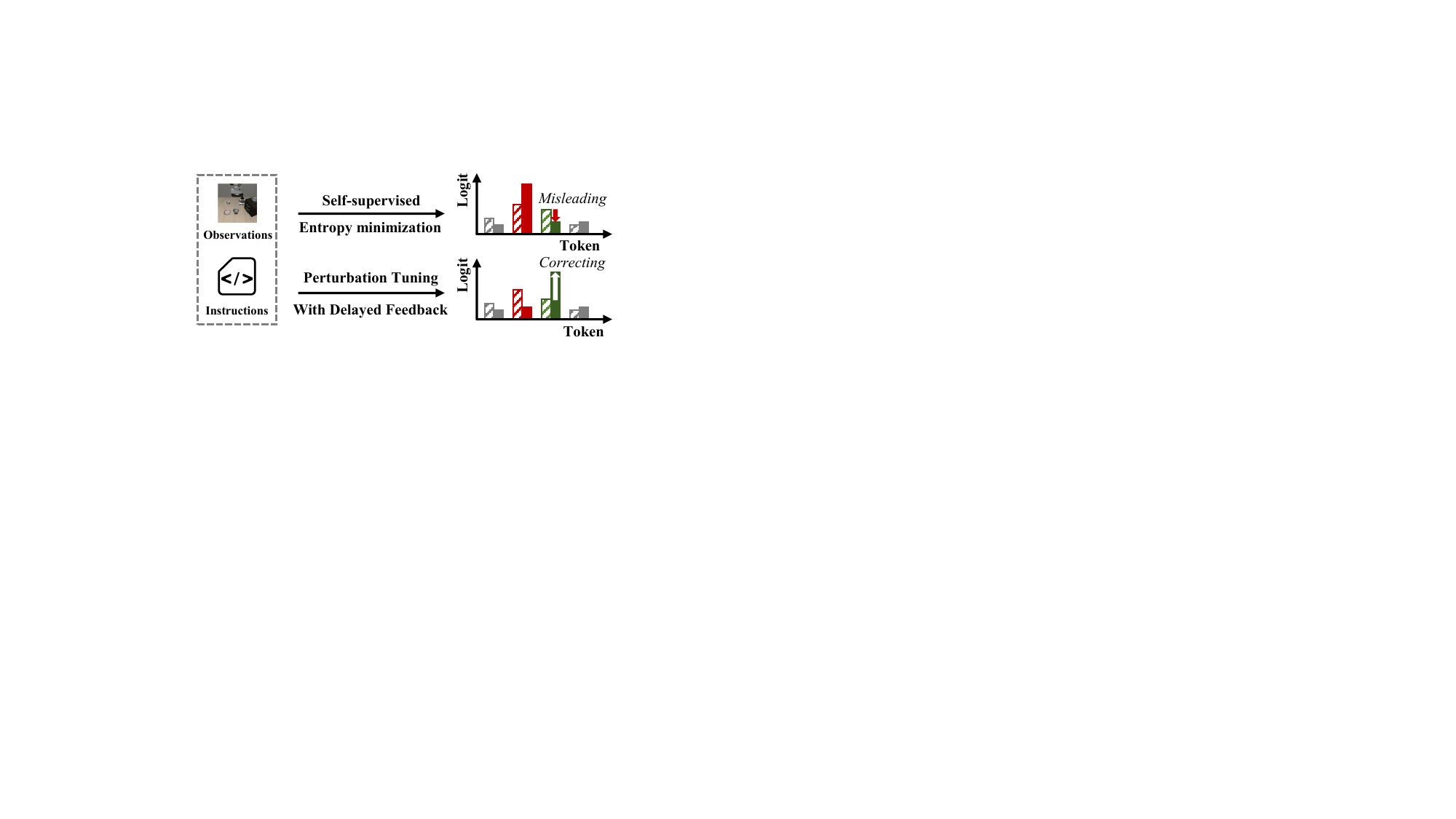}
    \caption{Comparison between traditional self-supervised test-time adaptation (TTA) and our PDF. Dashed bars indicate logits before adaptation; solid bars indicate logits after adaptation. Red bars denote incorrect-action logits; green bars denote correct-action logits. When VLAs become overconfident in incorrect behaviors, entropy-minimization-based TTA can amplify these errors by further boosting the wrong logits. In contrast, PDF corrects such misbehaviors by jointly mitigating the overconfidence issue and elevating the logits of correct actions, guiding the model toward accurate decisions.}
    \label{fig:comparision}
\end{figure}

One promising avenue to mitigate trajectory overfitting in VLAs is test-time adaptation (TTA)~\cite{scap,kff}, which allows models to refine their policies dynamically during inference. Among existing approaches, the most widely adopted are verifier-based TTA methods~\cite{robomonky,v-gps}, which employ a pretrained verifier to score candidate actions and select the most promising one via a best-of-$N$ voting scheme. Although effective, these methods incur substantial computational overhead from verifier pretraining and repeated rollouts, and they struggle to generalize to unseen environments with different dynamics or visual statistics. To address these limitations, recent efforts have explored verifier-free TTA strategies for VLAs~\cite{mgselect}, which adapt models in real time using only unlabeled test samples. However, without ground-truth feedback, these methods often suffer from adaptation instability or suboptimal accuracy under severe distribution shifts~\cite{bitta}. The main limitation lies in their reliance on self-supervised confidence metrics—such as entropy minimization~\cite{tent}—which may become \textit{unreliable} when model predictions are miscalibrated.

To overcome these limitations, we propose Perturbation learning with Delayed Feedback (PDF), a lightweight and verifier-free TTA framework designed to improve VLA robustness without parameter updates, which is demonstrated in Figure~\ref{fig:comparision}. PDF consists of two key components:
(1) Uncertainty-Based Action Voting, which adaptively allocates augmentation resources based on the model’s decision uncertainty, efficiently balancing decision performance and inference cost; and
(2) Delayed Feedback-Guided Adaptation, a logit perturbation module with delayed feedback, which refines the model’s action distribution using delayed feedback signals from the environment to correct overconfident decisions in a TTA manner.
Extensive experiments across multiple VLA benchmarks, including robotic manipulation LIBERO and visual control games Atari, demonstrate that PDF substantially improves test-time decision performance while maintaining real-time efficiency.

Our main contributions are summarized as follows:
\begin{enumerate}
    \item We systematically identify and analyze the phenomenon of trajectory overfitting, where VLAs rely on the spurious correlation between actions and entities, instead of semantic task grounding, revealing a fundamental cause of their fragility under environmental shifts.
    \item We propose PDF, a verifier-free TTA framework that enhances VLA decision performance through uncertainty-based action voting and delayed feedback-guided adaptation, while keeping the base model parameters frozen.
    \item Extensive experiments across diverse VLA benchmarks, including LIBERO and Atari, demonstrate that PDF consistently improves test-time decision performance.
\end{enumerate}

\begin{figure*}[t]
    \centering
    \includegraphics[width=1.0\linewidth]{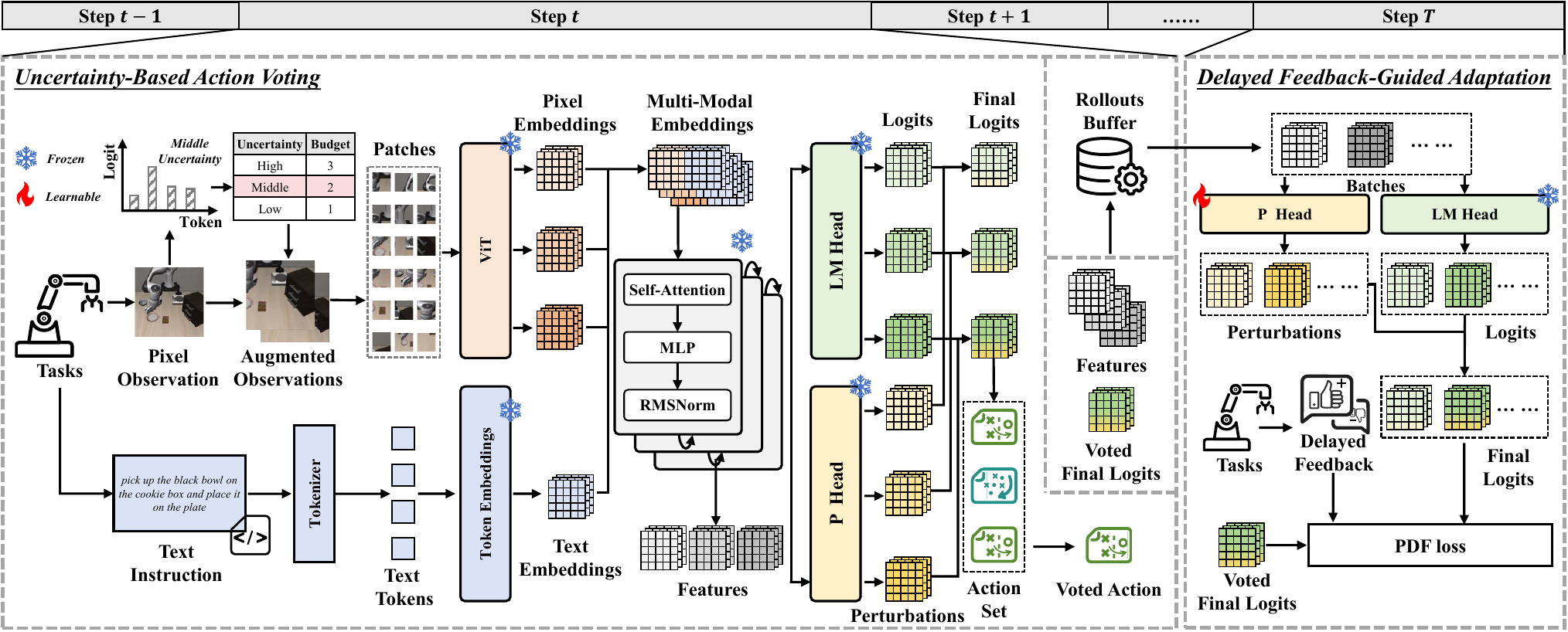}
    \caption{The overall framework of our proposed PDF. At test time, the VLA receives pixel observation $o_t$ and instruction $c_t$. Action\mbox{-}logit uncertainty $\mathcal{U}_t$ is estimated to allocate an adaptive augmentation budget $N_t$. Original and augmented observations are jointly encoded into multimodal features $f_t$, which are passed to an LM head to produce decision logits and to a perturbation head $h_\theta(\cdot)$ to generate logit perturbations. Final logits yield candidate actions, and the final action $a_t$ is selected by majority voting. The features and the voted final logits are stored in a rollout buffer $\mathcal{D}$. After each episode, a batch of features, voted final logits, and feedback is sampled to compute the PDF loss (Eq.~\ref{equ:pdf-loss}). Only the P head $h_\theta(\cdot)$ is updated, enabling efficient and stable adaptation while keeping all VLA parameters frozen.}
    \label{fig:framework}
\end{figure*}

\section{Related Works}
\subsection{Vision-Language-Action Models}
The strong open-world perception and reasoning capabilities of Vision-Language Models (VLMs) have driven their rapid adoption in robotic control tasks. Recently, Vision-Language-Action models (VLAs)~\cite{metavla,tracevla,vla1,vla2,CoT-VLA,rfm-1,LINGO-2} have emerged as an integrated and promising approach for bridging perception, language, and control. OpenVLA~\cite{openvla}, a prominent open-source framework, supports training VLA models by integrating vision, language, and robotic control, providing a scalable platform for embodied-AI research. Gato~\cite{gato}, developed by DeepMind, is a generalist model that unifies vision, language, and action tasks within a single architecture and demonstrates strong versatility across diverse modalities.  RT-1~\cite{RT-2} and RT-2~\cite{RT-2}, developed by Google, are robotic transformers for real-world manipulation tasks, with RT-2 extending RT-1 by incorporating language understanding to support more complex instruction following. Collectively, these works advance the integration of vision, language, and action in autonomous robotic agents.
% \subsection{Test-Time Adaptation} 
% TTA~\cite{tent, Delta, memo, tta1, DomainAdaptor, tta2, tta3, D3CTTA,causiam,scap,kff} has become a key paradigm for improving model robustness to distribution shifts between training and testing. Its core idea is to adapt models to new domains during inference without access to source data, thereby improving generalization in real-world scenarios. TENT~\cite{tent} introduces online adaptation by minimizing prediction entropy and updating batch-normalization statistics. Subsequent work refines batch-normalization updates to achieve more stable adaptation. For example, DIGA~\cite{DIGA} applies an exponential moving average to update normalization statistics, whereas TTN~\cite{TTN} modulates batch-normalization layers based on their sensitivity to domain shift, maintaining a balance between source and target distributions. Methods such as AugMix~\cite{augmix} and MEMO~\cite{memo} further enhance robustness by combining data augmentation with entropy minimization.

% Another line of work~\cite{tta5, tta6, GIPSO, PROGRAM, tta7, tta8, AdaNPC} explores pseudo-label-based adaptation strategies. To improve reliability, several methods~\cite{tta6, tta8, AdaNPC} use high-confidence samples to build prototype representations, reducing the influence of noisy labels and improving adaptation stability.

\subsection{Test-Time Adaptation in Vision-Language-Action Models}
To improve the robustness and adaptability of VLAs in real-world settings, TTA has been actively explored. FEEDTTA~\cite{FeedTTA} performs online adaptation for vision-language navigation via reinforcement learning with binary episodic feedback, but on a small navigation model. V-GPS~\cite{v-gps} reranks action candidates from a pretrained VLA using a learned value function, enabling test-time policy selection without fine-tuning. RoboMonkey~\cite{robomonky} scales at test time via sampling-and-verification with a VLM-based verifier. By contrast, SimpleVLA-RL~\cite{simplevla-rl} scales during training—rather than at test time—using reinforcement learning with outcome rewards. MG-Select~\cite{mgselect} estimates confidence via KL divergence between predicted and reference action distributions and selects the best action among samples without an external verifier. 

Our PDF instead improves confidence by mitigating the spurious correlation through data augmentation and refining the policy with delayed feedback, yielding a robust TTA scheme for VLAs.

% To improve the robustness and adaptability of VLAs in real-world settings, TTA has been actively explored. FEEDTTA~\cite{FeedTTA} introduces an online adaptation framework for vision-language navigation that updates the policy during testing via reinforcement learning with binary episodic feedback but with small navigation model. V-GPS~\cite{v-gps} uses a learned value function to rerank action candidates produced by a pretrained VLA, enabling test-time policy selection without fine-tuning. RoboMonkey~\cite{robomonky} performs test-time scaling via a sampling-and-verification mechanism, in which a VLM-based verifier selects the optimal action from multiple candidates. In contrast, SimpleVLA-RL~\cite{simplevla-rl} scales VLA capability during training—rather than only at test time—by using reinforcement learning with outcome rewards.

% MG-Select~\cite{mgselect} is conceptually related to our PDF and measures confidence using the KL divergence between predicted and reference action distributions, selecting the best action from multiple samples without an external verifier. In contrast, our PDF method improves confidence by mitigating spurious correlations through data augmentation and refining the policy using delayed feedback, ultimately yielding a  TTA scheme for VLAs.

\section{Methods}
In this section, we present Perturbation learning with Delayed Feedback (PDF), a plug-and-play TTA strategy designed for VLAs. Figure~\ref{fig:framework} depicts the overall architecture of our proposed PDF. The following subsections elaborate on each core component.
% \textbf{P}erturbation tuning with \textbf{D}elayed \textbf{F}eedback (PDF)

\subsection{Prelimiaries} \label{sec:pre}
We begin by formulating the VLAs problem as a Markov Decision Process (MDP), defined by the tuple $ M=(\mathcal{S}, \mathcal{A}, P, R) $, where $ \mathcal{S} \subseteq \mathbb{R}^n $ and $ \mathcal{A} \subseteq \mathbb{R}^m $ denote continuous state and action spaces, respectively. At each time step $ t $, the agent observes a multimodal state $ s_t=(o_t,c_t) \in \mathcal{S}$, consisting of a pixel observation $ o_t \in \mathcal{O} $ and a textual instruction $ c_t \in \mathcal{C} $. Given the current state $s_t$, the agent selects an action $ a_t \in \mathcal{A} $, causing the environment to transition to a new state $ s_{t+1} \sim P(\cdot\mid s_t,a_t) $. Upon task completion, the agent receives delayed feedback signals $ r_i \in \mathcal{R}, i \in [1, T] $, which encapsulate task completion status or reward information.

% Recently, VLAs~\cite{openvla,spatialvla,pi0} have demonstrated remarkable capabilities in sequential decision-making tasks, spanning robotic control scenarios and visually grounded control tasks. In robotic control tasks, such as those in the LIBERO~\cite{libero} benchmark, the state $ s_t \in \mathcal{S} $ typically comprises pixel observations from a fixed-viewpoint camera along with corresponding task instructions, providing both spatial and semantic context for decision-making. The agent’s action $a_t = [\Delta x_t, \Delta y_t, \Delta z_t, \Delta u_t, \Delta v_t, \Delta w_t, g_t]^\top$ specifies incremental translations and rotations of the end-effector, along with a binary gripper command controlling its open/close state. 
% In visually grounded control tasks, such as Atari~\cite{atari}, the state $s_t$ is typically represented by pixel observations encoding the spatial and temporal dynamics of the environment, often in the form of stacked RGB frames. The action space $\mathcal{A}$ is discrete, comprising a finite set of control commands (e.g., directional movements or key presses), denoted as $\mathcal{A}=\{a_1, a_2, \dots, a_K\}$. The delayed feedback signals $\mathcal{R}$ correspond to the cumulative returns obtained upon task completion or episode termination.

\subsection{Uncertainty-Based Action Voting}\label{sec:uav}
To improve the robustness of VLAs during test time, when VLAs interact with environments, the uncertainty of the model's decision is quantified using uncertainty estimation. Based on the estimated uncertainty, adaptive data augmentation is applied to mitigate overfitting and attention imbalance, thereby improving the reliability of action generation while preserving inference efficiency. 

Specifically, given a state $s_t = (o_t, c_t)$, the pixel observation $o_t$ is encoded by a visual encoder into pixel embeddings $e_{o_t}$, while the textual instruction $c_t$ is tokenized and embedded into text embeddings $e_{c_t}$. Then, both embeddings are concatenated and projected into features $f_t$. At last, the features are processed by an MLP $h_{\phi}(\cdot)$, referred to as the Large Model head (LM head), to generate the logits $z_t$ for each token. 

To evaluate the uncertainty of the model's decision under the current state, we estimate predictive uncertainty from the output logits, defined as the normalized Shannon entropy of the predicted action distribution:

\begin{equation} \label{equ:uncertainry}
\mathcal{U}_t = - \frac{1}{\log K} \sum_{k=1}^{K} p(a_k\mid s_k) \log p(a_k \mid s_k),
\end{equation}
where $K$ denotes the number of possible action tokens and $p(a_t = a_k \mid s_t) = \mathrm{softmax}(z_t)_k$. 

In sequential decision-making tasks, a single erroneous action can result in complete task failure, with high-uncertainty actions being particularly prone to causing such outcomes. To mitigate action-generation uncertainty, we employ a simple yet effective strategy—data augmentation. However, data augmentation introduces significant computational overhead and therefore should be applied selectively.
Therefore, augmented observations are generated by applying a set of transformations $\mathcal{T} = \{T_1, \dots, T_N\}$, where the augmentation budget is adaptively determined by the model’s uncertainty:
\begin{equation}
N_t = N_{\max} \cdot \mathcal{U}_t,
\end{equation}
where $N_{\max}$ denotes the maximum augmentation budget. Both the original observation $o_t$ and its augmented views $\{T_j(o_t)\}_{j=1}^{N_t}$ are then divided into patches and encoded into pixel embeddings. The resulting pixel embeddings are concatenated with text embeddings to form a unified multimodal sequence and processed by the causal transformer to produce features $f_t$. 

Finally, these features are projected by two parallel heads, the LM head $h_{\phi}(\cdot)$ and the Perturbation head (P head) $h_{\theta}(\cdot)$:
\begin{equation}\label{equ:perturbated-logtis}
\tilde{z}_t = h_{\phi}(f_t) + \lambda h_{\theta}(f_t),
\end{equation}
where $h_{\phi}(\cdot)$ generates the decision logits, $h_{\theta}(\cdot)$ outputs the learnable perturbations and $\lambda$ is a hyper-parameter. The resulting final logits $\tilde{z}_t$ are detokenized into a set of candidate actions, from which the final action $a_t$ is then selected through majority voting. The features of both the original and augmented observations are stored in a rollout buffer $\mathcal{D} $ for optimization once feedback is available.

It is important to note that all parameters of the VLA-including the visual encoder, 
token embedding layers, causal transformer, LM head and P head-remain frozen throughout this process.

\subsection{Delayed Feedback-Guided Adaptation}\label{sec:dfa}
Upon an episode completion, the agent receives delayed feedback $r \in \mathcal{R}$, which reflects overall task performance, such as task success, failure, or cumulative reward. A batch of features $f_b$ is then sampled from the rollout buffer $\mathcal{D}$, and the final logits $\tilde{z}_b$ with gradients are computed as
\begin{equation}
\tilde{z}_b = h_{\phi}(f_b) + \lambda h_{\theta}(f_b).
\end{equation}

Then, the policy $\tilde{\pi}_b$ is generated by applying the softmax function to the final logits $\tilde{z}_b$. And the P head is subsequently optimized by
\begin{equation} \label{equ:pdf-loss}
\mathcal{L}_{\mathrm{PDF}}=-(r-b)\log\pi_\phi+\lambda_\mathrm{KL}\mathbb{I}[r>b]\operatorname{KL}(\pi_\phi\parallel\tilde{\pi}),
\end{equation}
where the first term is a REINFORCE-style objective increasing the likelihood of actions from the perturbed policy $\tilde{\pi}(s_t)$ when feedback $r$ exceeds the baseline $b$. A KL regularizer is gated by $\mathbb{I}[r>b]$, applied only under positive feedback to stabilize updates and accelerate convergence while allowing flexible exploration when $r \le b$. Perturbations are thus steered toward successful action directions and discouraged otherwise. Only the P head $h_\theta$ is updated; all other VLA parameters remain frozen for efficient, stable test-time adaptation.

\begin{table*}[t]
\centering
\caption{Performance comparison on the LIBERO benchmark (Spatial, Object, Goal, Long). $\dagger$ denotes reproduced results and SR indicates Success Rate (\%). Blue numbers indicate SOTA within each task suite.}
\label{tab:exp-libero}
\setlength{\tabcolsep}{6pt}
\renewcommand{\arraystretch}{1.2}
\begin{adjustbox}{max width=\textwidth}
\begin{tabular}{lcccccccccccccc}
\toprule
\multirow{2}{*}{\textbf{Method}} &
\multirow{2}{*}{\textbf{Pub.}} & 
\multirow{2}{*}{\textbf{Param.}} & 
\multicolumn{2}{c}{\textbf{Spatial}} &
\multicolumn{2}{c}{\textbf{Object}} &
\multicolumn{2}{c}{\textbf{Goal}} &
\multicolumn{2}{c}{\textbf{Long}} &
\multicolumn{2}{c}{\textbf{Avg.}}\\
% \cmidrule(lr){3-4} \cmidrule(lr){5-6} \cmidrule(lr){7-8} \cmidrule(lr){9-10} \cmidrule(lr){11-12}
\cmidrule(lr){4-5} \cmidrule(lr){6-7} \cmidrule(lr){8-9} \cmidrule(lr){10-11} \cmidrule(lr){12-13}
% \rowcolor{lightgray}
&  &  & SR $\uparrow$ & Rank $\downarrow$ & SR $\uparrow$ & Rank $\downarrow$ & SR $\uparrow$ & Rank $\downarrow$ & SR $\uparrow$ & Rank $\downarrow$ & SR $\uparrow$ & Rank $\downarrow$ \\
\midrule
PackNet~\cite{packnet}           & CVPR’18     &    - & 0.63 & 11 & 0.60 & 10 & 0.75 &  8 & 0.25 & 12 & \cellcolor{yellow!20}0.56 & \cellcolor{yellow!20}10.2 \\
ER~\cite{er}                     & Arxiv’19    &    - & 0.56 & 12 & 0.44 & 12 & 0.49 & 11 & 0.32 & 11 & \cellcolor{yellow!20}0.45 & \cellcolor{yellow!20}11.5 \\
SeqL~\cite{libero}               & NeurIPS’23  &    - & 0.20 & 13 & 0.26 & 13 & 0.22 & 12 & 0.15 & 13 & \cellcolor{yellow!20}0.21 & \cellcolor{yellow!20}12.8 \\
MTL~\cite{libero}                & NeurIPS’23  &    - & 0.83 &  6 & 0.54 & 11 & 0.80 &  3 & 0.48 &  9 & \cellcolor{yellow!20}0.66 & \cellcolor{yellow!20}7.2 \\
\midrule
ATM~\cite{atm}                   & RSS’23      &    - & 0.69 & 10 & 0.68 &  8 & 0.78 &  5 & 0.39 & 10 & \cellcolor{yellow!20}0.63 & \cellcolor{yellow!20}10.5 \\
OpenVLA~\cite{openvla}           & CoRL’24     &    - & 0.79 &  8 & 0.86 &  4 & 0.85 &  2 & 0.51 &  7 & \cellcolor{yellow!20}0.75 & \cellcolor{yellow!20}5.2 \\
OpenVLA$^\dagger$~\cite{openvla} & CoRL’24     &    - & 0.85 &  2 & 0.64 &  9 & 0.76 &  6 & 0.53 &  6 & \cellcolor{yellow!20}0.69 & \cellcolor{yellow!20}5.8 \\
DP~\cite{dp}                     & IJRR’25     &    - & 0.78 &  9 & \textbf{\textcolor{bestblue}{0.92}} & \textbf{\textcolor{bestblue}{1}} & 0.68 & 10 & 0.51 & 8 & \cellcolor{yellow!20}0.72 & \cellcolor{yellow!20}7 \\
\midrule
OCTO~\cite{octo}                 & RSS’24      &  93M & 0.79 & 8 & 0.86 & 4 & 0.85 & 2 & 0.51 & 7 & \cellcolor{yellow!20}0.75 & \cellcolor{yellow!20}5.3 \\
TraceVLA~\cite{tracevla}         & ICLR’25     & 130M & 0.85 & 4 & 0.85 & 5 & 0.75 & 7 & 0.54 & 4 & \cellcolor{yellow!20}0.75 & \cellcolor{yellow!20}6.5 \\
OpenVLA-DPO~\cite{tgrpo}         & Arxiv’25    & 130M & 0.84 & 5 & 0.89 & 2 & 0.79 & 4 & 0.53 & 5 & \cellcolor{yellow!20}0.76 & \cellcolor{yellow!20}4 \\
SFT-4LIBERO~\cite{metavla}       & Arxiv’25    & 130M & 0.85 & 3 & 0.87 & 3 & 0.77 & 5 & 0.55 & 3 & \cellcolor{yellow!20}0.76 & \cellcolor{yellow!20}3.5 \\
\midrule
MG-Select~\cite{mgselect}        & Arxiv’25    & 130M & 0.82 & 7 & 0.73 & 6 & 0.73 & 9 & 0.55 & 2 & \cellcolor{yellow!20}0.71 & \cellcolor{yellow!20}6 \\
\rowcolor{lightgreen}
\textbf{PDF (Ours)}              & -           &   9M & 
                                                        \textbf{\textcolor{bestblue}{0.90}} & \textbf{\textcolor{bestblue}{1}} &
                                                        0.72 & 7 &
                                                        \textbf{\textcolor{bestblue}{0.86}} & \textbf{\textcolor{bestblue}{1}} &
                                                        \textbf{\textcolor{bestblue}{0.59}} & \textbf{\textcolor{bestblue}{1}} &
                                                        \textbf{\textcolor{bestblue}{0.77}} & \textbf{\textcolor{bestblue}{2.5}} \\
\bottomrule
\end{tabular}
\end{adjustbox}
\end{table*}

\begin{figure*}[t]
    \centering
    \includegraphics[width=1.0\linewidth]{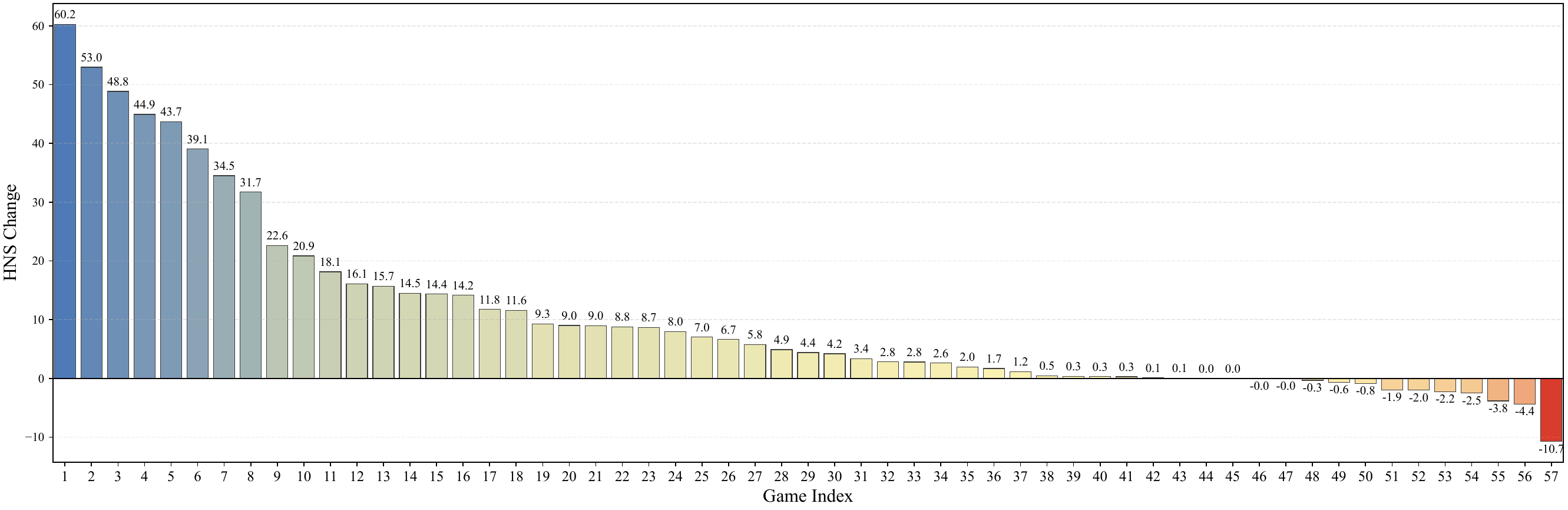}
    \caption{Human normalized score changes across 57 Atari games. Blue bars show performance improvements, orange bars indicate degradation. Games are sorted by improvement magnitude, with BOXING (+60.25\%) showing maximum gain and BATTLE ZONE (-10.72\%) showing maximum decline. 47/57 games demonstrate positive performance changes, with 11.28\% mean improvement.}
    \label{fig:atari-change}
\end{figure*}

\section{Experiments}
To evaluate the effectiveness of the proposed PDF, we conduct experiments across diverse decision-making simulation environments. Specifically, we employ OpenVLA \cite{openvla}, a VLA fine-tuned on the LIBERO \cite{libero} benchmark and Jat \cite{jat}—an open-source implementation of GATO \cite{gato}—for the Atari suite \cite{atari}. This combination of environments and baselines enables a comprehensive evaluation of the superiority of our method across diverse task domains.

\subsection{Experimental Setup}
LIBERO~\cite{libero}: We use four 10‑task suites (Spatial, Object, Goal, Long) in the LIBERO Franka Panda simulation with RGB views, robot states, task text, and delta end‑effector actions. Each task: 50 eval rollouts; metric is success rate. Policies are trained for 50 episodes with OpenVLA~\cite{openvla} before evaluation.

Atari~\cite{atari}: We evaluate all 57 Atari 2600 games (pixel input, 4–18 discrete actions). Episodes end on life loss, level completion, or at 108{,}000 frames. Each game: 50 eval episodes; metric is Human‑Normalized Score $(HNS = (\mathrm{Score}-\mathrm{RandomScore})/(\mathrm{HumanScore}-\mathrm{RandomScore}))$. Base model Jat~\cite{jat} is trained 50 episodes per game. All LIBERO and Atari experiments run on a single Tesla V100‑PCIE‑32GB.

\subsection{Performance Comparison Across Benchmarks}
To demonstrate the effectiveness of PDF, we conduct extensive experiments across multiple benchmarks.

Table \ref{tab:exp-libero} summarizes results on four LIBERO suites. PDF attains the best average performance (0.77 SR, 2.5 mean rank), exceeding prior VLA variants and adaptation baselines. It reaches 0.89 on Spatial (vs. 0.85 for reproduced OpenVLA) and 0.86 on Goal, slightly ahead of OCTO and OpenVLA. Gains are largest on Long-horizon tasks (0.59, +4.1 over the best baseline), indicating stronger sequential decision quality. Compared to MG-Select, PDF improves average SR by nearly 6 points, showing the advantage of uncertainty-weighted action voting and delayed-feedback perturbation learning over sample filtering. PDF also matches or surpasses strong pretrained models (TraceVLA, SFT-4LIBERO) without any fine-tuning. These results underscore PDF’s effectiveness in mitigating trajectory overfitting and boosting robustness in spatially complex, long-horizon settings.

PDF is also evaluated on Atari-57, attaining 1.07 HNS vs. 0.97 for the Jat base (+0.10). Figure~\ref{fig:atari-change} shows per-game HNS deltas: 47/57 games improve. Largest gains appear in BOXING (+0.60), TIME PILOT (+0.53), ATLANTIS (+0.49), and ASSAULT (+0.45). Our method uses 9M trainable parameters and requires no gradient access to the base model, whereas baselines use 93–130M parameters and require base‑model gradients.

\subsection{Impact of Data Augmentation and Feedbacks}

\begin{table*}[t]
    \centering
    \caption{Performance comparison of OpenVLA and PDF variants (with and without Delayed Feedback (DF) and Data Augmentation (DA)) across LIBERO benchmark suites (Spatial, Object, Goal, and Long). Reported as success rates over 10 tasks with average performance. 
    % Bold indicates the best per task.
    }
    % \resizebox{\textwidth}{!}{
    \begin{tabular}{llccccccccccccc}
    \toprule
    \textbf{Suite} & \textbf{Method} & 0 & 1 & 2 & 3 & 4 & 5 & 6 & 7 & 8 & 9 & \textbf{Avg.} \\
    \midrule
    \multirow{5}{*}{\rotatebox{90}{\textbf{Spatial}}} 
        & OpenVLA     & 0.94 & 0.94 & 0.86 & 0.96 & 0.64 & 0.88 & 0.92 & 0.86 & 0.74 & 0.76 & \cellcolor{yellow!20}0.85 \\
        & PDF w/o DF  & 0.78 & 0.84 & 0.90 & 0.90 & 0.78 & 0.68 & 0.92 & 0.70 & 0.80 & 0.74 & \cellcolor{yellow!20}0.80 \\
        & PDF w/o DA  & 0.94 & 0.94 & 0.88 & 0.94 & 0.68 & 0.88 & 0.92 & 0.88 & 0.76 & 0.76 & \cellcolor{yellow!20}0.86 \\
        &PDF          & 0.94 & 0.96 & 0.98 & 0.98 & 0.80 & 0.88 & 0.92 & 0.88 & 0.80 & 0.80 & \cellcolor{yellow!20}0.89 \\
        &\cellcolor{lightgreen}$\Delta\uparrow$ & \cellcolor{lightgreen}0.00 & \cellcolor{lightgreen}0.02 & \cellcolor{lightgreen}0.12 & \cellcolor{lightgreen}0.02 & \cellcolor{lightgreen}0.16 & \cellcolor{lightgreen}0.00 & \cellcolor{lightgreen}0.00 & \cellcolor{lightgreen}0.02 & \cellcolor{lightgreen}0.06 & \cellcolor{lightgreen}0.04 & \cellcolor{lightgreen}0.04 \\
    \midrule
    \multirow{5}{*}{\rotatebox{90}{\textbf{Object}}} 
        & OpenVLA     & 0.64 & 0.68 & 0.78 & 0.34 & 0.90 & 0.66 & 0.50 & 0.62 & 0.36 & 0.74 & \cellcolor{yellow!20}0.62 \\
        & PDF w/o DF  & 0.86 & 0.50 & 0.74 & 0.62 & 0.68 & 0.16 & 0.10 & 0.28 & 0.36 & 0.74 & \cellcolor{yellow!20}0.50 \\
        & PDF w/o DA  & 0.88 & 0.68 & 0.82 & 0.40 & 0.94 & 0.70 & 0.48 & 0.66 & 0.36 & 0.76 & \cellcolor{yellow!20}0.67 \\
        &PDF          & 0.96 & 0.74 & 0.82 & 0.66 & 0.96 & 0.66 & 0.56 & 0.66 & 0.40 & 0.80 & \cellcolor{yellow!20}0.72 \\
        &\cellcolor{lightgreen}$\Delta\uparrow$ & \cellcolor{lightgreen}0.32 & \cellcolor{lightgreen}0.06 & \cellcolor{lightgreen}0.04 & \cellcolor{lightgreen}0.32 & \cellcolor{lightgreen}0.06 & \cellcolor{lightgreen}0.00 & \cellcolor{lightgreen}0.06 & \cellcolor{lightgreen}0.04 & \cellcolor{lightgreen}0.04 & \cellcolor{lightgreen}0.06 & \cellcolor{lightgreen}0.10\\
    \midrule
    \multirow{5}{*}{\rotatebox{90}{\textbf{Goal}}} 
        & OpenVLA     & 0.62 & 0.90 & 0.84 & 0.66 & 0.94 & 0.90 & 0.78 & 1.00 & 0.86 & 0.68 & \cellcolor{yellow!20}0.81 \\
        & PDF w/o DF  & 0.70 & 0.96 & 0.88 & 0.34 & 0.94 & 0.78 & 0.82 & 0.98 & 0.68 & 0.62 & \cellcolor{yellow!20}0.77 \\
        & PDF w/o DA  & 0.62 & 0.94 & 0.86 & 0.60 & 0.92 & 0.94 & 0.80 & 1.00 & 0.76 & 0.66 & \cellcolor{yellow!20}0.81 \\
        &PDF          & 0.72 & 0.98 & 0.90 & 0.66 & 0.96 & 0.96 & 0.86 & 1.00 & 0.82 & 0.70 & \cellcolor{yellow!20}0.85 \\
        &\cellcolor{lightgreen}$\Delta\uparrow$ & \cellcolor{lightgreen}0.10 & \cellcolor{lightgreen}0.08 & \cellcolor{lightgreen}0.06 & \cellcolor{lightgreen}0.00 & \cellcolor{lightgreen}0.02 & \cellcolor{lightgreen}0.06 & \cellcolor{lightgreen}0.08 & \cellcolor{lightgreen}0.00 & \cellcolor{lightgreen}-0.04 & \cellcolor{lightgreen}0.02 & \cellcolor{lightgreen}0.04\\
    \midrule
    \multirow{4}{*}{\rotatebox{90}{\textbf{Long}}} 
        & OpenVLA     & 0.62 & 0.64 & 0.58 & 0.50 & 0.44 & 0.86 & 0.50 & 0.66 & 0.20 & 0.58 & \cellcolor{yellow!20}0.56 \\
        & PDF w/o DF  & 0.70 & 0.64 & 0.58 & 0.54 & 0.52 & 0.82 & 0.48 & 0.62 & 0.20 & 0.52 & \cellcolor{yellow!20}0.56 \\
        & PDF w/o DA  & 0.72 & 0.60 & 0.58 & 0.50 & 0.46 & 0.88 & 0.50 & 0.66 & 0.20 & 0.56 & \cellcolor{yellow!20}0.57 \\
        & PDF         & 0.72 & 0.64 & 0.58 & 0.56 & 0.56 & 0.88 & 0.50 & 0.68 & 0.20 & 0.58 & \cellcolor{yellow!20}0.59 \\
        &\cellcolor{lightgreen}$\Delta\uparrow$ & \cellcolor{lightgreen}0.10 & \cellcolor{lightgreen}0.00 & \cellcolor{lightgreen}0.00 & \cellcolor{lightgreen}0.06 & \cellcolor{lightgreen}0.12 & \cellcolor{lightgreen}0.02 & \cellcolor{lightgreen}0.00 & \cellcolor{lightgreen}0.02 & \cellcolor{lightgreen}0.00 & \cellcolor{lightgreen}0.00 & \cellcolor{lightgreen}0.03\\
    \bottomrule
    \end{tabular}
    \label{tab:ab-module}
\end{table*}

To evaluate the impact of the data augmentation (DA) mechanism and perturbation learning with delayed feedback (DF), we conduct extensive experiments on the four LIBERO suites. Note that “PDF w/o DF” denotes OpenVLA that selects actions solely via multi‑view voting (without delayed feedback), and “PDF w/o DA” denotes OpenVLA that learns perturbations using only the original pixel observations (without data augmentation). Table~\ref{tab:ab-module} presents an ablation comparing the baseline OpenVLA with variants of the proposed PDF across the LIBERO suites. Results show that integrating both DF and DA in PDF is effective. Overall, PDF attains the highest average success rates across all four suites. This superiority holds for most tasks within each suite, as indicated by boldface entries in the table. 

Moreover, the ablation reveals the contribution of each component. The PDF w/o DA variant surpasses the OpenVLA baseline on most tasks. By contrast, the PDF w/o DF variant shows a substantial drop across suites, with success rates dropping to 0.50 on Object and 0.77 on Goal. These results suggest that DF is critical for robustness, particularly in object‑manipulation and goal‑conditioned tasks. The synergy between DF and DA is evident. The full PDF model outperforms both variants and yields more stable results across suites, confirming that both DF and DA are integral to the observed performance gains.

\subsection{Ablation Study on Data Augmentation Budget}
\begin{figure}
    \centering
    \includegraphics[width=0.95\linewidth]{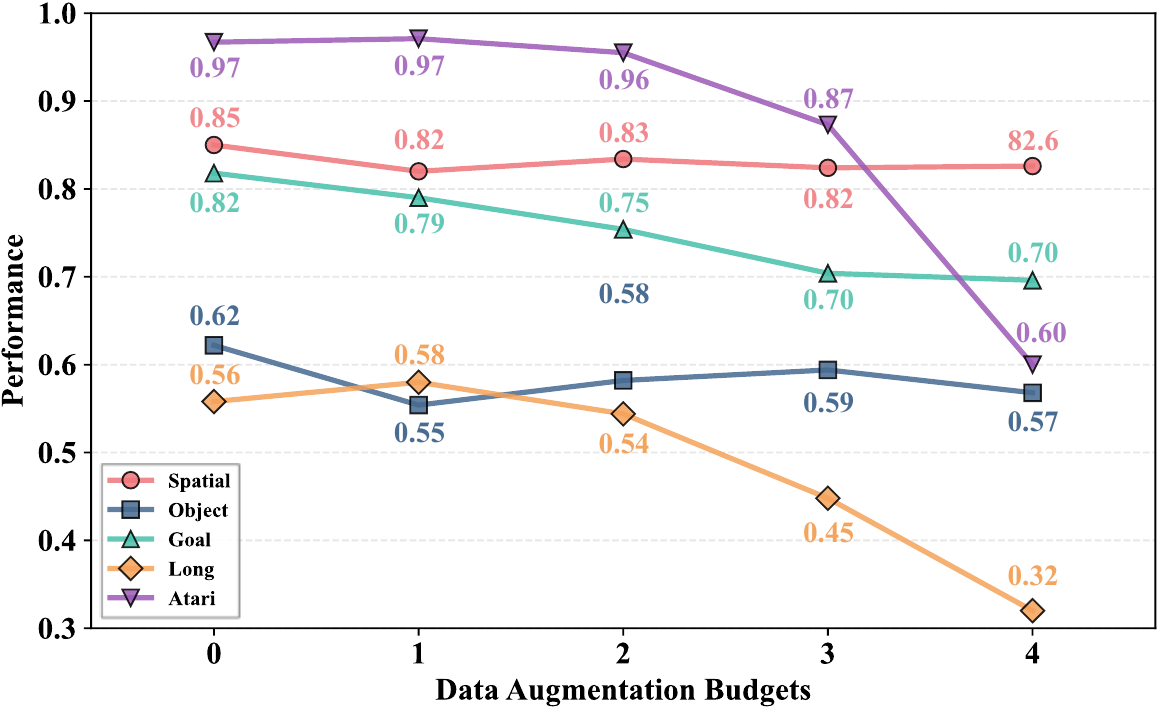}
    \caption{Performance degradation under increasing data augmentation budgets. All five benchmarks (LIBERO-Spatial, LIBERO-Object, LIBERO-Goal, LIBERO-Long, and Atari) show performance decline as augmentation budget increases from 0 to 4, indicating that excessive data augmentation harms model effectiveness.}
    \label{fig:data-augmentation-budget}
\end{figure}
Given that data augmentation (DA) is a crucial module in PDF and its budget affects performance, we evaluate its impact across the four LIBERO suites and the Atari benchmark. Figure~\ref{fig:data-augmentation-budget} summarizes performance under five augmentation budgets (0-4, higher uncertainty indicates more data augmentations) for the LIBERO and Atari benchmarks. As the test-time augmentation budget (i.e., the number of augmented views sampled for voting) increases, performance generally declines across benchmarks. Increasing the data augmentation budget in PDF generally degrades performance across LIBERO and Atari benchmarks due to noise accumulation, with a small budget (max of 3) offering the best trade-off to mitigate trajectory overfitting without introducing low-quality views. Adaptive, benchmark-specific budget ceilings are recommended for optimal results in structure-dependent tasks. Overall, we set the max data augmentation budget to 3, because moderate augmentation is benign, whereas aggressive budgets degrade structure‑dependent tasks; adaptive, benchmark‑specific ceilings are preferable to uniform maximization.

\begin{table}[t]
    \centering
    \caption{Performance comparison of dim-wise voting and action-wise voting across Atari and LIBERO tasks.}
    \begin{adjustbox}{max width=\linewidth}
    \begin{tabular}{lccc}
        \toprule
        \textbf{Task} & \textbf{Dim-wise} & \textbf{Action-wise} \\
        \midrule 
        Alien         & \textbf{\textcolor{bestblue}{0.26}}  &  -\\ 
        Black Bowl    & 0.68              & \textbf{\textcolor{bestblue}{0.88}} \\
        Alphabet Soup & \textbf{\textcolor{bestblue}{0.86}} & 0.64  \\
        Bowl on Stove & \textbf{\textcolor{bestblue}{0.96}} & 0.90 \\
        Moka Pot      & \textbf{\textcolor{bestblue}{0.58}} & \textbf{\textcolor{bestblue}{0.58}} \\
        \bottomrule        
    \end{tabular}
    \end{adjustbox}
    \label{tab:ab-voting}
\end{table}

\subsection{Ablation Study on Loss Functions}
\begin{figure}[t]
    \centering
    \includegraphics[width=1.0\linewidth]{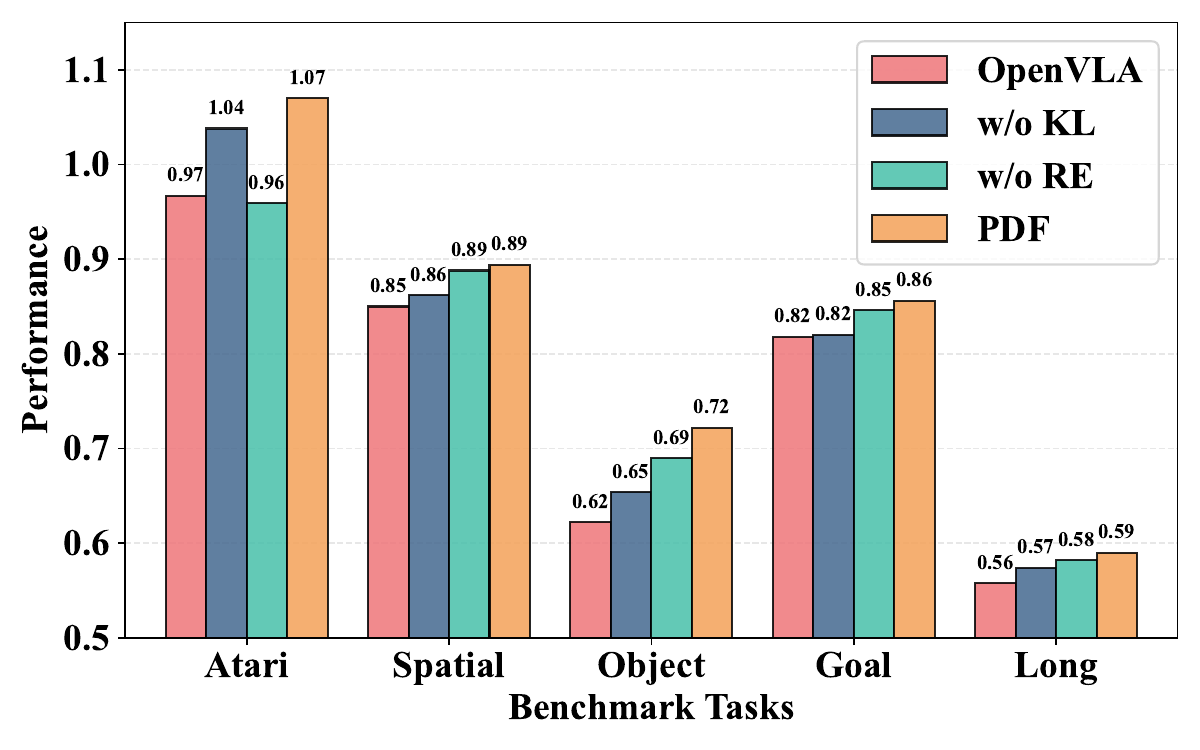}
    \caption{Performance comparison across five benchmarks shows that both the KL divergence and the REINFORCE-style term contribute to the full PDF model's superior performance. Removing either component results in degraded performance, with KL divergence showing greater impact on final results.}
    \label{fig:ab-loss}
\end{figure}

The PDF objective comprises two components: a REINFORCE‑style policy gradient term and a KL regularization term. To evaluate the contribution of each term, we conduct experiments on the four LIBERO suites and the Atari benchmark. Figure~\ref{fig:ab-loss} compares: (i) the base model OpenVLA, (ii) a variant without the KL term (w/o KL), (iii) a variant without the REINFORCE‑style term (w/o RE), and (iv) the full PDF objective. The full objective yields the highest scores; the largest absolute/relative gains occur on Atari (+0.1) and LIBERO‑Object (+0.1). Removing the KL term markedly boosts Atari (+0.07) but yields only modest gains on other suites; removing the REINFORCE‑style term slightly harms Atari (-0.01) yet improves LIBERO‑Object (+0.07) and shows smaller positive shifts elsewhere. Moderate improvements are observed on LIBERO‑Spatial (+0.04) and LIBERO‑Goal (+0.04), while LIBERO‑Long shows a smaller gain (+0.03). These patterns suggest KL term and the REINFORCE‑style term provide complementary benefits—stability and calibrated exploration vs. reward‑aligned action refinement—whose combination yields additive, non‑redundant gains.

\subsection{Comparison between Voting Mechanisms}
To evaluate voting strategies, we compare three variants. Dimension-wise voting takes a majority vote per action dimension, while action-wise voting votes over full action tuples. As shown in Table~\ref{tab:ab-voting}, on Atari (Alien) the action is effectively one-dimensional, making dimension-wise and action-wise voting equivalent. On LIBERO, dimension-wise voting best mitigates trajectory overfitting by allowing cross-view agreement on individual action components, enabling more flexible deviations from the original policy. Action-wise voting often falls back to the baseline because views rarely agree on the entire action tuple, yielding mixed results. Overall, dimension-wise voting provides the most reliable robustness–plasticity trade-off.

\subsection{Visualization and Case Study}
\begin{figure*}[t]
    \centering
    \includegraphics[width=0.83\linewidth]{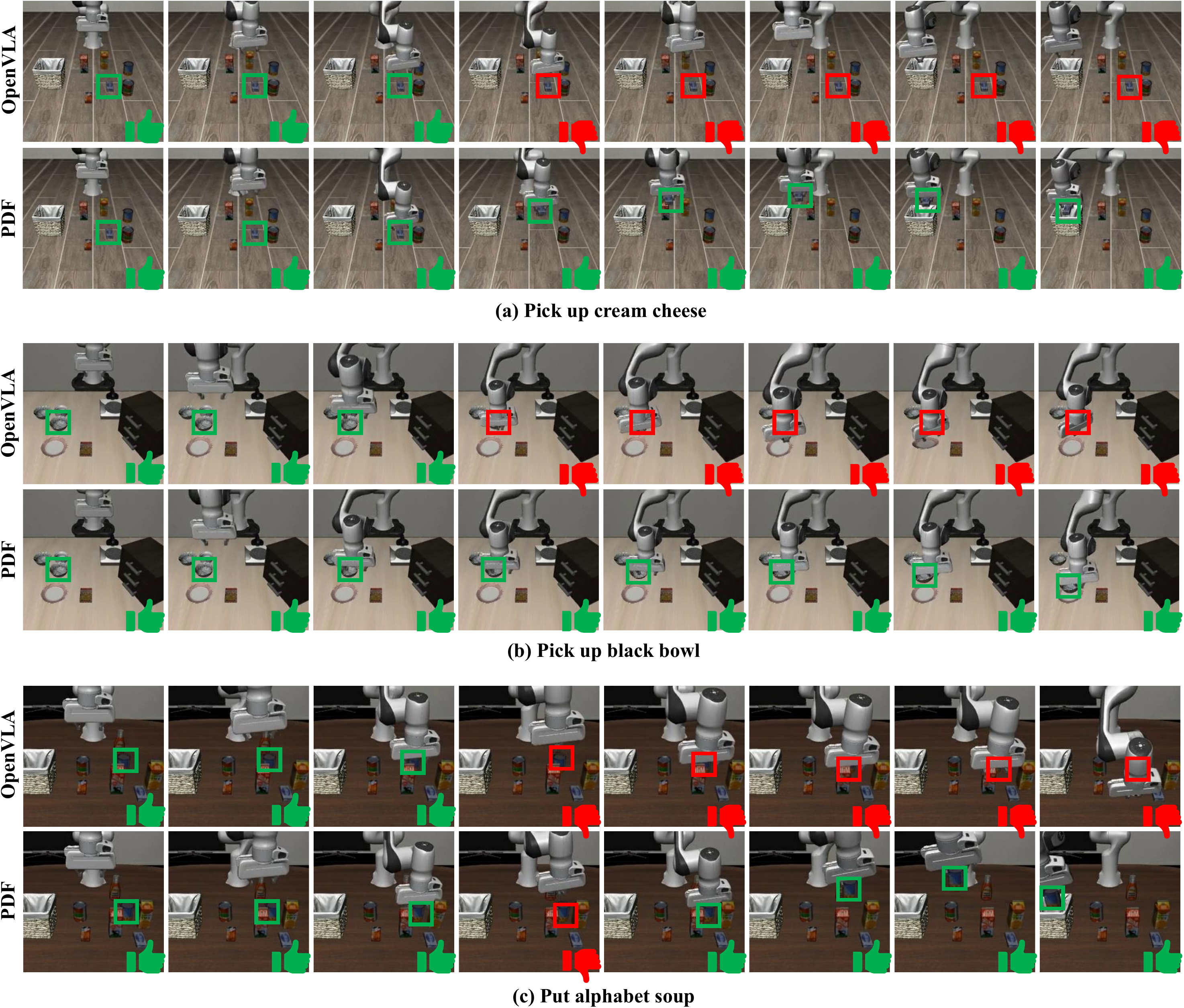}
    \caption{Visual comparison of OpenVLA and PDF on three tasks. The green thumb indicates that the agent performed the correct action, while the red thumb indicates an incorrect action. The box represents the target entity, with green indicating that the target entity has been operated correctly and red indicating incorrect operation.}
    \label{fig:case-study}
\end{figure*}

To further assess the effectiveness of PDF, we conduct qualitative case studies on three LIBERO tasks: (i) “pick up the cream cheese and place it in the basket” (\textit{Pick up cream cheese}), (ii) “pick up the black bowl on the wooden cabinet and place it on the plate” (\textit{Pick up black bowl}), and (iii) “put both the alphabet soup and the tomato sauce in the basket” (\textit{Put alphabet soup}). As shown in Figure~\ref{fig:case-study}(a,b), OpenVLA exhibits trajectory overfitting: it executes a demonstration-like motion plan but disregards whether the target object has been successfully grasped, leading to failure to complete the tasks. In contrast, PDF manifests a more robust, state-aware policy under the same conditions and reliably completes both \textit{Pick up cream cheese} and \textit{Pick up black bowl}. Interestingly, on \textit{Put alphabet soup} (Figure~\ref{fig:case-study}(c)), PDF initially fails to secure the alphabet soup but promptly reattempts and succeeds, indicating decision-making rather than open-loop replay. Overall, these case studies highlight PDF’s resilience to demonstration bias and its capacity to leverage feedback for corrective actions, resulting in higher task completion reliability.

\section{Conclusions and Limitations}
% We present PDF, a TTA method that enhances the decision performance of VLAs in sequential decision-making tasks. Our analysis shows that VLAs exhibit trajectory overfitting: models overemphasize spurious correlation and memorized action sequences, which makes them fragile under subtle environmental shifts. PDF addresses this issue via two components: uncertainty-based action voting and delayed feedback-guided adaptation. Evaluated on robotic manipulation and 3D visual control benchmarks, PDF consistently outperforms strong VLA baselines and alternative test-time adaptation methods, delivering substantial gains in robustness and task success with minimal computational overhead. These results demonstrate that PDF is a practical and efficient solution for improving the reliability of decision-making agents in dynamic environments, enabling more adaptable and resilient VLAs without costly retraining. However, PDF does not fully solve trajectory overfitting; rather, it alleviates its negative impact. In future work, we will investigate the underlying mechanisms of this phenomenon to address the challenge more fundamentally.

We introduce PDF, a verifier-free TTA method that improves VLA decision performance in sequential decision making.VLAs suffer from trajectory overfitting—overweighting the spurious correlation and replaying memorized action chains—reducing reliability under subtle environmental shifts. PDF addresses this via (i) uncertainty-based action voting and (ii) delayed-feedback–guided adaptation. On LIBERO robotic manipulation and Atari-57, PDF surpasses vanilla VLAs and VLAs with TTA baselines, improving  decision performance with minimal overhead. However, PDF does not fully solve trajectory overfitting; rather, it alleviates its negative impact. In future work, we will address the challenge more fundamentally.

\section*{Acknowlegedements}
This work was a research achievement of National Engineering Research Center of New Electronic Publishing Technologies, and was supported by the National Natural Science Foundation of China (62406313, 62376011), the Postdoctoral Fellowship Program of China Postdoctoral Science Foundation, Grant No. YJB20250283, and the National Key R\&D Program of China (2024YFA1410000).

% This work is supported by National Natural Science Foundation of China No. 62406313, Postdoctoral Fellowship Program of China Postdoctoral Science Foundation, Grant No. YJB20250283, National Natural Science Foundation of China (62376011) and the National Key R\&D Program of China (2024YFA1410000). And this work is a research achievement of National Engineering Research Center of New Electronic Publishing Technologies.

{
    \small
    \bibliographystyle{ieeenat_fullname}
    \bibliography{main}

@article{kff,
  author       = {Jiahuan Zhou and
                  Chao Zhu and
                  Zhenyu Cui and
                  Zichen Liu and
                  Xu Zou and
                  Gang Hua},
  title        = {Class-aware Domain Knowledge Fusion and Fission for Continual Test-Time
                  Adaptation},
  journal      = {CoRR},
  volume       = {abs/2510.12150},
  year         = {2025},
  url          = {https://doi.org/10.48550/arXiv.2510.12150},
  doi          = {10.48550/ARXIV.2510.12150},
  eprinttype    = {arXiv},
  eprint       = {2510.12150},
  bibsource    = {dblp computer science bibliography, https://dblp.org}
}

@inproceedings{scap,
  author       = {Chenyu Zhang and
                  Kunlun Xu and
                  Zichen Liu and
                  Yuxin Peng and
                  Jiahuan Zhou},
  title        = {{SCAP:} Transductive Test-Time Adaptation via Supportive Clique-based
                  Attribute Prompting},
  booktitle    = {{IEEE/CVF} Conference on Computer Vision and Pattern Recognition,
                  {CVPR} 2025, Nashville, TN, USA, June 11-15, 2025},
  pages        = {30032--30041},
  publisher    = {Computer Vision Foundation / {IEEE}},
  year         = {2025},
  url          = {https://openaccess.thecvf.com/content/CVPR2025/html/Zhang\_SCAP\_Transductive\_Test-Time\_Adaptation\_via\_Supportive\_Clique-based\_Attribute\_Prompting\_CVPR\_2025\_paper.html},
  doi          = {10.1109/CVPR52734.2025.02795},
  bibsource    = {dblp computer science bibliography, https://dblp.org}
}

@article{bitta,
  author       = {Taeckyung Lee and
                  Sorn Chottananurak and
                  Junsu Kim and
                  Jinwoo Shin and
                  Taesik Gong and
                  Sung{-}Ju Lee},
  title        = {Test-Time Adaptation with Binary Feedback},
  journal      = {CoRR},
  volume       = {abs/2505.18514},
  year         = {2025},
  url          = {https://doi.org/10.48550/arXiv.2505.18514},
  doi          = {10.48550/ARXIV.2505.18514},
  eprinttype    = {arXiv},
  eprint       = {2505.18514},
  bibsource    = {dblp computer science bibliography, https://dblp.org}
}

@inproceedings{jarvisvla,
  author       = {Muyao Li and
                  Zihao Wang and
                  Kaichen He and
                  Xiaojian Ma and
                  Yitao Liang},
  editor       = {Wanxiang Che and
                  Joyce Nabende and
                  Ekaterina Shutova and
                  Mohammad Taher Pilehvar},
  title        = {{JARVIS-VLA:} Post-Training Large-Scale Vision Language Models to
                  Play Visual Games with Keyboards and Mouse},
  booktitle    = {Findings of the Association for Computational Linguistics, {ACL} 2025,
                  Vienna, Austria, July 27 - August 1, 2025},
  pages        = {17878--17899},
  publisher    = {Association for Computational Linguistics},
  year         = {2025},
  url          = {https://aclanthology.org/2025.findings-acl.920/},
  bibsource    = {dblp computer science bibliography, https://dblp.org}
}

@inproceedings{octo,
  author       = {Dibya Ghosh and
                  Homer Rich Walke and
                  Karl Pertsch and
                  Kevin Black and
                  Oier Mees and
                  Sudeep Dasari and
                  Joey Hejna and
                  Tobias Kreiman and
                  Charles Xu and
                  Jianlan Luo and
                  You Liang Tan and
                  Lawrence Yunliang Chen and
                  Quan Vuong and
                  Ted Xiao and
                  Pannag R. Sanketi and
                  Dorsa Sadigh and
                  Chelsea Finn and
                  Sergey Levine},
  editor       = {Dana Kulic and
                  Gentiane Venture and
                  Kostas E. Bekris and
                  Enrique Coronado},
  title        = {Octo: An Open-Source Generalist Robot Policy},
  booktitle    = {Robotics: Science and Systems XX, Delft, The Netherlands, July 15-19,
                  2024},
  year         = {2024},
  url          = {https://doi.org/10.15607/RSS.2024.XX.090},
  doi          = {10.15607/RSS.2024.XX.090},
  bibsource    = {dblp computer science bibliography, https://dblp.org}
}

@article{metavla,
  author       = {Chen Li and
                  Zhantao Yang and
                  Han Zhang and
                  Fangyi Chen and
                  Chenchen Zhu and
                  Anudeepsekhar Bolimera and
                  Marios Savvides},
  title        = {MetaVLA: Unified Meta Co-training For Efficient Embodied Adaption},
  journal      = {CoRR},
  volume       = {abs/2510.05580},
  year         = {2025},
  url          = {https://doi.org/10.48550/arXiv.2510.05580},
  doi          = {10.48550/ARXIV.2510.05580},
  eprinttype    = {arXiv},
  eprint       = {2510.05580},
  bibsource    = {dblp computer science bibliography, https://dblp.org}
}

@article{tgrpo,
  author       = {Zengjue Chen and
                  Runliang Niu and
                  He Kong and
                  Qi Wang},
  title        = {{TGRPO} :Fine-tuning Vision-Language-Action Model via Trajectory-wise
                  Group Relative Policy Optimization},
  journal      = {CoRR},
  volume       = {abs/2506.08440},
  year         = {2025},
  url          = {https://doi.org/10.48550/arXiv.2506.08440},
  doi          = {10.48550/ARXIV.2506.08440},
  eprinttype    = {arXiv},
  eprint       = {2506.08440},
  bibsource    = {dblp computer science bibliography, https://dblp.org}
}

@inproceedings{tracevla,
  author       = {Ruijie Zheng and
                  Yongyuan Liang and
                  Shuaiyi Huang and
                  Jianfeng Gao and
                  Hal Daum{\'{e}} III and
                  Andrey Kolobov and
                  Furong Huang and
                  Jianwei Yang},
  title        = {TraceVLA: Visual Trace Prompting Enhances Spatial-Temporal Awareness
                  for Generalist Robotic Policies},
  booktitle    = {The Thirteenth International Conference on Learning Representations,
                  {ICLR} 2025, Singapore, April 24-28, 2025},
  publisher    = {OpenReview.net},
  year         = {2025},
  url          = {https://openreview.net/forum?id=b1CVu9l5GO},
  bibsource    = {dblp computer science bibliography, https://dblp.org}
}

@inproceedings{atm,
  author       = {Chuan Wen and
                  Xingyu Lin and
                  John Ian Reyes So and
                  Kai Chen and
                  Qi Dou and
                  Yang Gao and
                  Pieter Abbeel},
  editor       = {Dana Kulic and
                  Gentiane Venture and
                  Kostas E. Bekris and
                  Enrique Coronado},
  title        = {Any-point Trajectory Modeling for Policy Learning},
  booktitle    = {Robotics: Science and Systems XX, Delft, The Netherlands, July 15-19,
                  2024},
  year         = {2024},
  url          = {https://doi.org/10.15607/RSS.2024.XX.092},
  doi          = {10.15607/RSS.2024.XX.092},
  bibsource    = {dblp computer science bibliography, https://dblp.org}
}

@article{dp,
  author       = {Cheng Chi and
                  Zhenjia Xu and
                  Siyuan Feng and
                  Eric Cousineau and
                  Yilun Du and
                  Benjamin Burchfiel and
                  Russ Tedrake and
                  Shuran Song},
  title        = {Diffusion policy: Visuomotor policy learning via action diffusion},
  journal      = {Int. J. Robotics Res.},
  volume       = {44},
  number       = {10-11},
  pages        = {1684--1704},
  year         = {2025},
  url          = {https://doi.org/10.1177/02783649241273668},
  doi          = {10.1177/02783649241273668},
  bibsource    = {dblp computer science bibliography, https://dblp.org}
}

@inproceedings{packnet,
  author       = {Arun Mallya and
                  Svetlana Lazebnik},
  title        = {PackNet: Adding Multiple Tasks to a Single Network by Iterative Pruning},
  booktitle    = {2018 {IEEE} Conference on Computer Vision and Pattern Recognition,
                  {CVPR} 2018, Salt Lake City, UT, USA, June 18-22, 2018},
  pages        = {7765--7773},
  publisher    = {Computer Vision Foundation / {IEEE} Computer Society},
  year         = {2018},
  url          = {http://openaccess.thecvf.com/content\_cvpr\_2018/html/Mallya\_PackNet\_Adding\_Multiple\_CVPR\_2018\_paper.html},
  doi          = {10.1109/CVPR.2018.00810},
  bibsource    = {dblp computer science bibliography, https://dblp.org}
}

@misc{er,
      title={On Tiny Episodic Memories in Continual Learning}, 
      author={Arslan Chaudhry and Marcus Rohrbach and Mohamed Elhoseiny and Thalaiyasingam Ajanthan and Puneet K. Dokania and Philip H. S. Torr and Marc'Aurelio Ranzato},
      year={2019},
      eprint={1902.10486},
      archivePrefix={arXiv},
      primaryClass={cs.LG},
      url={https://arxiv.org/abs/1902.10486}, 
}

@article{openvla-oft,
  author       = {Moo Jin Kim and
                  Chelsea Finn and
                  Percy Liang},
  title        = {Fine-Tuning Vision-Language-Action Models: Optimizing Speed and Success},
  journal      = {CoRR},
  volume       = {abs/2502.19645},
  year         = {2025},
  url          = {https://doi.org/10.48550/arXiv.2502.19645},
  doi          = {10.48550/ARXIV.2502.19645},
  eprinttype    = {arXiv},
  eprint       = {2502.19645},
  bibsource    = {dblp computer science bibliography, https://dblp.org}
}

@article{robomonky,
  author       = {Jacky Kwok and
                  Christopher Agia and
                  Rohan Sinha and
                  Matthew Foutter and
                  Shulu Li and
                  Ion Stoica and
                  Azalia Mirhoseini and
                  Marco Pavone},
  title        = {RoboMonkey: Scaling Test-Time Sampling and Verification for Vision-Language-Action
                  Models},
  journal      = {CoRR},
  volume       = {abs/2506.17811},
  year         = {2025},
  url          = {https://doi.org/10.48550/arXiv.2506.17811},
  doi          = {10.48550/ARXIV.2506.17811},
  eprinttype    = {arXiv},
  eprint       = {2506.17811},
  bibsource    = {dblp computer science bibliography, https://dblp.org}
}

@article{jat,
  author       = {Quentin Gallou{\'{e}}dec and
                  Edward Beeching and
                  Cl{\'{e}}ment Romac and
                  Emmanuel Dellandr{\'{e}}a},
  title        = {Jack of All Trades, Master of Some, a Multi-Purpose Transformer Agent},
  journal      = {CoRR},
  volume       = {abs/2402.09844},
  year         = {2024},
  url          = {https://doi.org/10.48550/arXiv.2402.09844},
  doi          = {10.48550/ARXIV.2402.09844},
  eprinttype    = {arXiv},
  eprint       = {2402.09844},
  bibsource    = {dblp computer science bibliography, https://dblp.org}
}

@article{gato,
  author       = {Scott E. Reed and
                  Konrad Zolna and
                  Emilio Parisotto and
                  Sergio G{\'{o}}mez Colmenarejo and
                  Alexander Novikov and
                  Gabriel Barth{-}Maron and
                  Mai Gimenez and
                  Yury Sulsky and
                  Jackie Kay and
                  Jost Tobias Springenberg and
                  Tom Eccles and
                  Jake Bruce and
                  Ali Razavi and
                  Ashley Edwards and
                  Nicolas Heess and
                  Yutian Chen and
                  Raia Hadsell and
                  Oriol Vinyals and
                  Mahyar Bordbar and
                  Nando de Freitas},
  title        = {A Generalist Agent},
  journal      = {Trans. Mach. Learn. Res.},
  volume       = {2022},
  year         = {2022},
  url          = {https://openreview.net/forum?id=1ikK0kHjvj},
  bibsource    = {dblp computer science bibliography, https://dblp.org}
}

@inproceedings{openvla,
  author       = {Moo Jin Kim and
                  Karl Pertsch and
                  Siddharth Karamcheti and
                  Ted Xiao and
                  Ashwin Balakrishna and
                  Suraj Nair and
                  Rafael Rafailov and
                  Ethan Paul Foster and
                  Pannag R. Sanketi and
                  Quan Vuong and
                  Thomas Kollar and
                  Benjamin Burchfiel and
                  Russ Tedrake and
                  Dorsa Sadigh and
                  Sergey Levine and
                  Percy Liang and
                  Chelsea Finn},
  editor       = {Pulkit Agrawal and
                  Oliver Kroemer and
                  Wolfram Burgard},
  title        = {OpenVLA: An Open-Source Vision-Language-Action Model},
  booktitle    = {Conference on Robot Learning, 6-9 November 2024, Munich, Germany},
  series       = {Proceedings of Machine Learning Research},
  volume       = {270},
  pages        = {2679--2713},
  publisher    = {{PMLR}},
  year         = {2024},
  url          = {https://proceedings.mlr.press/v270/kim25c.html},
  bibsource    = {dblp computer science bibliography, https://dblp.org}
}

@article{pi0,
  author       = {Kevin Black and
                  Noah Brown and
                  Danny Driess and
                  Adnan Esmail and
                  Michael Equi and
                  Chelsea Finn and
                  Niccolo Fusai and
                  Lachy Groom and
                  Karol Hausman and
                  Brian Ichter and
                  Szymon Jakubczak and
                  Tim Jones and
                  Liyiming Ke and
                  Sergey Levine and
                  Adrian Li{-}Bell and
                  Mohith Mothukuri and
                  Suraj Nair and
                  Karl Pertsch and
                  Lucy Xiaoyang Shi and
                  James Tanner and
                  Quan Vuong and
                  Anna Walling and
                  Haohuan Wang and
                  Ury Zhilinsky},
  title        = {{\(\pi\)}\({}_{\mbox{0}}\): {A} Vision-Language-Action Flow Model
                  for General Robot Control},
  journal      = {CoRR},
  volume       = {abs/2410.24164},
  year         = {2024},
  url          = {https://doi.org/10.48550/arXiv.2410.24164},
  doi          = {10.48550/ARXIV.2410.24164},
  eprinttype    = {arXiv},
  eprint       = {2410.24164},
  bibsource    = {dblp computer science bibliography, https://dblp.org}
}

@article{spatialvla,
  author       = {Delin Qu and
                  Haoming Song and
                  Qizhi Chen and
                  Yuanqi Yao and
                  Xinyi Ye and
                  Yan Ding and
                  Zhigang Wang and
                  JiaYuan Gu and
                  Bin Zhao and
                  Dong Wang and
                  Xuelong Li},
  title        = {SpatialVLA: Exploring Spatial Representations for Visual-Language-Action
                  Model},
  journal      = {CoRR},
  volume       = {abs/2501.15830},
  year         = {2025},
  url          = {https://doi.org/10.48550/arXiv.2501.15830},
  doi          = {10.48550/ARXIV.2501.15830},
  eprinttype    = {arXiv},
  eprint       = {2501.15830},
  bibsource    = {dblp computer science bibliography, https://dblp.org}
}

@inproceedings{tent,
  author       = {Dequan Wang and
                  Evan Shelhamer and
                  Shaoteng Liu and
                  Bruno A. Olshausen and
                  Trevor Darrell},
  title        = {Tent: Fully Test-Time Adaptation by Entropy Minimization},
  booktitle    = {9th International Conference on Learning Representations, {ICLR} 2021,
                  Virtual Event, Austria, May 3-7, 2021},
  publisher    = {OpenReview.net},
  year         = {2021},
  url          = {https://openreview.net/forum?id=uXl3bZLkr3c},
  bibsource    = {dblp computer science bibliography, https://dblp.org}
}

@article{simplevla-rl,
  author       = {Haozhan Li and
                  Yuxin Zuo and
                  Jiale Yu and
                  Yuhao Zhang and
                  Zhaohui Yang and
                  Kaiyan Zhang and
                  Xuekai Zhu and
                  Yuchen Zhang and
                  Tianxing Chen and
                  Ganqu Cui and
                  Dehui Wang and
                  Dingxiang Luo and
                  Yuchen Fan and
                  Youbang Sun and
                  Jia Zeng and
                  Jiangmiao Pang and
                  Shanghang Zhang and
                  Yu Wang and
                  Yao Mu and
                  Bowen Zhou and
                  Ning Ding},
  title        = {SimpleVLA-RL: Scaling {VLA} Training via Reinforcement Learning},
  journal      = {CoRR},
  volume       = {abs/2509.09674},
  year         = {2025},
  url          = {https://doi.org/10.48550/arXiv.2509.09674},
  doi          = {10.48550/ARXIV.2509.09674},
  eprinttype    = {arXiv},
  eprint       = {2509.09674},
  bibsource    = {dblp computer science bibliography, https://dblp.org}
}

@inproceedings{v-gps,
  author       = {Mitsuhiko Nakamoto and
                  Oier Mees and
                  Aviral Kumar and
                  Sergey Levine},
  editor       = {Pulkit Agrawal and
                  Oliver Kroemer and
                  Wolfram Burgard},
  title        = {Steering Your Generalists: Improving Robotic Foundation Models via
                  Value Guidance},
  booktitle    = {Conference on Robot Learning, 6-9 November 2024, Munich, Germany},
  series       = {Proceedings of Machine Learning Research},
  volume       = {270},
  pages        = {4996--5013},
  publisher    = {{PMLR}},
  year         = {2024},
  url          = {https://proceedings.mlr.press/v270/nakamoto25a.html},
  bibsource    = {dblp computer science bibliography, https://dblp.org}
}

@inproceedings{FeedTTA,
title={Test-Time Adaptation for Online Vision-Language Navigation with Feedback-based Reinforcement Learning},
author={Sungjune Kim and Gyeongrok Oh and Heeju Ko and Daehyun Ji and Dongwook Lee and Byung-Jun Lee and Sujin Jang and Sangpil Kim},
booktitle={Forty-second International Conference on Machine Learning},
year={2025},
url={https://openreview.net/forum?id=K4GaB4fdIq}
}

@inproceedings{RT-2,
  author       = {Brianna Zitkovich and
                  Tianhe Yu and
                  Sichun Xu and
                  Peng Xu and
                  Ted Xiao and
                  Fei Xia and
                  Jialin Wu and
                  Paul Wohlhart and
                  Stefan Welker and
                  Ayzaan Wahid and
                  Quan Vuong and
                  Vincent Vanhoucke and
                  Huong T. Tran and
                  Radu Soricut and
                  Anikait Singh and
                  Jaspiar Singh and
                  Pierre Sermanet and
                  Pannag R. Sanketi and
                  Grecia Salazar and
                  Michael S. Ryoo and
                  Krista Reymann and
                  Kanishka Rao and
                  Karl Pertsch and
                  Igor Mordatch and
                  Henryk Michalewski and
                  Yao Lu and
                  Sergey Levine and
                  Lisa Lee and
                  Tsang{-}Wei Edward Lee and
                  Isabel Leal and
                  Yuheng Kuang and
                  Dmitry Kalashnikov and
                  Ryan Julian and
                  Nikhil J. Joshi and
                  Alex Irpan and
                  Brian Ichter and
                  Jasmine Hsu and
                  Alexander Herzog and
                  Karol Hausman and
                  Keerthana Gopalakrishnan and
                  Chuyuan Fu and
                  Pete Florence and
                  Chelsea Finn and
                  Kumar Avinava Dubey and
                  Danny Driess and
                  Tianli Ding and
                  Krzysztof Marcin Choromanski and
                  Xi Chen and
                  Yevgen Chebotar and
                  Justice Carbajal and
                  Noah Brown and
                  Anthony Brohan and
                  Montserrat Gonzalez Arenas and
                  Kehang Han},
  editor       = {Jie Tan and
                  Marc Toussaint and
                  Kourosh Darvish},
  title        = {{RT-2:} Vision-Language-Action Models Transfer Web Knowledge to Robotic
                  Control},
  booktitle    = {Conference on Robot Learning, CoRL 2023, 6-9 November 2023, Atlanta,
                  GA, {USA}},
  series       = {Proceedings of Machine Learning Research},
  volume       = {229},
  pages        = {2165--2183},
  publisher    = {{PMLR}},
  year         = {2023},
  url          = {https://proceedings.mlr.press/v229/zitkovich23a.html},
  bibsource    = {dblp computer science bibliography, https://dblp.org}
}

@inproceedings{vla1,
  author       = {Xinghang Li and
                  Minghuan Liu and
                  Hanbo Zhang and
                  Cunjun Yu and
                  Jie Xu and
                  Hongtao Wu and
                  Chilam Cheang and
                  Ya Jing and
                  Weinan Zhang and
                  Huaping Liu and
                  Hang Li and
                  Tao Kong},
  title        = {Vision-Language Foundation Models as Effective Robot Imitators},
  booktitle    = {The Twelfth International Conference on Learning Representations,
                  {ICLR} 2024, Vienna, Austria, May 7-11, 2024},
  publisher    = {OpenReview.net},
  year         = {2024},
  url          = {https://openreview.net/forum?id=lFYj0oibGR},
  bibsource    = {dblp computer science bibliography, https://dblp.org}
}

@inproceedings{3D-VLA,
  author       = {Haoyu Zhen and
                  Xiaowen Qiu and
                  Peihao Chen and
                  Jincheng Yang and
                  Xin Yan and
                  Yilun Du and
                  Yining Hong and
                  Chuang Gan},
  title        = {3D-VLA: {A} 3D Vision-Language-Action Generative World Model},
  booktitle    = {Forty-first International Conference on Machine Learning, {ICML} 2024,
                  Vienna, Austria, July 21-27, 2024},
  publisher    = {OpenReview.net},
  year         = {2024},
  url          = {https://openreview.net/forum?id=EZcFK8HupF},
  bibsource    = {dblp computer science bibliography, https://dblp.org}
}

@inproceedings{vla2,
  author       = {Jiangyong Huang and
                  Silong Yong and
                  Xiaojian Ma and
                  Xiongkun Linghu and
                  Puhao Li and
                  Yan Wang and
                  Qing Li and
                  Song{-}Chun Zhu and
                  Baoxiong Jia and
                  Siyuan Huang},
  title        = {An Embodied Generalist Agent in 3D World},
  booktitle    = {Forty-first International Conference on Machine Learning, {ICML} 2024,
                  Vienna, Austria, July 21-27, 2024},
  publisher    = {OpenReview.net},
  year         = {2024},
  url          = {https://openreview.net/forum?id=V4qV08Vk6S},
  bibsource    = {dblp computer science bibliography, https://dblp.org}
}

@inproceedings{CoT-VLA,
  author       = {Qingqing Zhao and
                  Yao Lu and
                  Moo Jin Kim and
                  Zipeng Fu and
                  Zhuoyang Zhang and
                  Yecheng Wu and
                  Zhaoshuo Li and
                  Qianli Ma and
                  Song Han and
                  Chelsea Finn and
                  Ankur Handa and
                  Tsung{-}Yi Lin and
                  Gordon Wetzstein and
                  Ming{-}Yu Liu and
                  Donglai Xiang},
  title        = {CoT-VLA: Visual Chain-of-Thought Reasoning for Vision-Language-Action
                  Models},
  booktitle    = {{IEEE/CVF} Conference on Computer Vision and Pattern Recognition,
                  {CVPR} 2025, Nashville, TN, USA, June 11-15, 2025},
  pages        = {1702--1713},
  publisher    = {Computer Vision Foundation / {IEEE}},
  year         = {2025},
  url          = {https://openaccess.thecvf.com/content/CVPR2025/html/Zhao\_CoT-VLA\_Visual\_Chain-of-Thought\_Reasoning\_for\_Vision-Language-Action\_Models\_CVPR\_2025\_paper.html},
  doi          = {10.1109/CVPR52734.2025.00166},
  bibsource    = {dblp computer science bibliography, https://dblp.org}
}

@misc{rfm-1,
  title={Introducing rfm-1: Giving robots human-like reason-ing capabilities},
  author={Sohn, A and Nagabandi, A and Florensa, C and Adelberg, D and Wu, D and Farooq, H and Clavera, I and Welborn, J and Chen, J and Mishra, N and others},
  year={2024}
}

@misc{LINGO-2,
  title={LINGO-2: Driving with Natural Language},
  author={Waywe Research Team and others},
  year={2024}
}

@article{LIBERO-Plus,
  title={LIBERO-Plus: In-depth Robustness Analysis of Vision-Language-Action Models},
  author={Fei, Senyu and Wang, Siyin and Shi, Junhao and Dai, Zihao and Cai, Jikun and Qian, Pengfang and Ji, Li and He, Xinzhe and Zhang, Shiduo and Fei, Zhaoye and others},
  journal={arXiv preprint arXiv:2510.13626},
  year={2025}
}

@article{atari,
  author       = {Marc G. Bellemare and
                  Yavar Naddaf and
                  Joel Veness and
                  Michael Bowling},
  title        = {The Arcade Learning Environment: An Evaluation Platform for General
                  Agents},
  journal      = {J. Artif. Intell. Res.},
  volume       = {47},
  pages        = {253--279},
  year         = {2013},
  url          = {https://doi.org/10.1613/jair.3912},
  doi          = {10.1613/JAIR.3912},
  bibsource    = {dblp computer science bibliography, https://dblp.org}
}

@inproceedings{libero,
  author       = {Bo Liu and
                  Yifeng Zhu and
                  Chongkai Gao and
                  Yihao Feng and
                  Qiang Liu and
                  Yuke Zhu and
                  Peter Stone},
  editor       = {Alice Oh and
                  Tristan Naumann and
                  Amir Globerson and
                  Kate Saenko and
                  Moritz Hardt and
                  Sergey Levine},
  title        = {{LIBERO:} Benchmarking Knowledge Transfer for Lifelong Robot Learning},
  booktitle    = {Advances in Neural Information Processing Systems 36: Annual Conference
                  on Neural Information Processing Systems 2023, NeurIPS 2023, New Orleans,
                  LA, USA, December 10 - 16, 2023},
  year         = {2023},
  url          = {http://papers.nips.cc/paper\_files/paper/2023/hash/8c3c666820ea055a77726d66fc7d447f-Abstract-Datasets\_and\_Benchmarks.html},
  bibsource    = {dblp computer science bibliography, https://dblp.org}
}

@article{mgselect,
  author       = {Suhyeok Jang and
                  Dongyoung Kim and
                  Changyeon Kim and
                  Youngsuk Kim and
                  Jinwoo Shin},
  title        = {Verifier-free Test-Time Sampling for Vision Language Action Models},
  journal      = {CoRR},
  volume       = {abs/2510.05681},
  year         = {2025},
  url          = {https://doi.org/10.48550/arXiv.2510.05681},
  doi          = {10.48550/ARXIV.2510.05681},
  eprinttype    = {arXiv},
  eprint       = {2510.05681},
  bibsource    = {dblp computer science bibliography, https://dblp.org}
}

@String(CVPR= {IEEE Conf. Comput. Vis. Pattern Recog.})

@String(ICLR = {Int. Conf. Learn. Represent.})

@String(CVPR  = {CVPR})

@String(ICLR  = {ICLR})
}


\clearpage
\setcounter{page}{1}
\maketitlesupplementary

\section{Psydo-Code}
We provide a brief overview of the training pipeline outlined in Algorithm~\ref{alg:pdf}, which implements PDF.

\begin{algorithm}
\caption{Perturbation Learning with Delayed Feedback (PDF)}
\label{alg:pdf}
\begin{algorithmic}[1]
\REQUIRE Pretrained VLA parameters $\phi$ (frozen); Perturbation head parameters $\theta$ (trainable); 
maximum augmentation budget $N_{\max}$; buffer $\mathcal{D}$.
\FOR{each episode}
    \FOR{each timestep $t$}
        \STATE Observe current state $s_t = (o_t, c_t)$
        \STATE Embed $o_t$ and $c_t$ to obtain multimodal feature $f_t$
        \STATE Compute logits $z_t = h_{\phi}(f_t)$
        \STATE Estimate uncertainty by Equation~\ref{equ:uncertainry}
        \STATE Determine augmentation budget $N_t = N_{\max}\cdot \mathcal{U}_t$
        \STATE Generate augmented views $\{T_j(o_t)\}_{j=1}^{N_t}$ 
        \FOR{each view $T_j(o_t)$}
            \STATE Encode to feature $f_{t,j}$
            \STATE Compute perturbated logits by Equation 3
        \ENDFOR
        \STATE Select final action $a_t$ by majority voting over $\tilde{z}_{t,j}$
        \STATE Interact with environment using $a_t$
        \STATE Store $\{f_{t,j}\}$ in buffer $\mathcal{D}$
    \ENDFOR
    \STATE Receive delayed feedback $r$ after episode
    \STATE Sample batch of features $f_b$ from $\mathcal{D}$
    \STATE Compute perturbed logits $\tilde{z}_b = h_{\phi}(f_b) + h_{\theta}(f_b)$
    \STATE Generate policy $\tilde{\pi}_b = \mathrm{softmax}(\tilde{z}_b)$
    \STATE Compute PDF loss by Equation 5
    \STATE Update perturbation head $\theta \leftarrow \theta - \eta\nabla_\theta \mathcal{L}_{\mathrm{PDF}}$
\ENDFOR
\end{algorithmic}
\end{algorithm}

\section{Additional Experiments Results on Atari 57}
Table \ref{tab:exp-jat-extend} presents the detailed results of PDF and JAT on the full Atari 57 benchmark.

\begin{table*}[t]
    \centering
    \begin{adjustbox}{width=\textwidth,center}
    \small
    \begin{tabular}{lccccc}
    \toprule
        Games        & ID & JAT (Raw Score) & JAT (Human Normalized Score)       & \textbf{PDF (Raw Score)} & \textbf{PDF (Human Normalized Score)} \\
    \midrule
    ALIEN & 22 & 1427.9 ± 540.28 & 0.17 & 2034.4 ± 560.47 & 0.26 \\
    AMIDAR & 34 & 105 ± 76.93 & 0.06 & 150.2 ± 57.05 & 0.08 \\
    ASSAULT & 4 & 1627.57 ± 799.09 & 2.7 & 1861 ± 742.69 & 3.15 \\
    ASTERIX & 51 & 861.5 ± 517.97 & 0.08 & 700 ± 361.25 & 0.06 \\
    ASTEROIDS & 42 & 1362 ± 461.03 & 0.01 & 1405 ± 241.92 & 0.01 \\
    ATLANTIS & 3 & 47047 ± 102148.27 & 2.11 & 54950 ± 12497.3 & 2.6 \\
    BANKHEIST & 38 & 972.6 ± 139.62 & 1.3 & 976 ± 109.01 & 1.3 \\
    BATTLEZONE & 57 & 17660 ± 6516.47 & 0.47 & 13700 ± 5348.83 & 0.36 \\
    BEAMRIDER & 48 & 757.76 ± 297.41 & 0.02 & 704.4 ± 250.38 & 0.02 \\
    BERZERK & 23 & 692.3 ± 337.1 & 0.23 & 910 ± 160.93 & 0.31 \\
    BOWLING & 37 & 22.48 ± 5.71 & 0 & 24.1 ± 5.72 & 0.01 \\
    BOXING & 1 & 92.77 ± 12.05 & 7.72 & 100 ± 22.36 & 8.33 \\
    BREAKOUT & 52 & 8.77 ± 5.81 & 0.25 & 8.2 ± 4.53 & 0.23 \\
    CENTIPEDE & 24 & 5807.4 ± 2289.32 & 0.37 & 6599.8 ± 3203.28 & 0.45 \\
    CHOPPERCOMMAND & 10 & 2448 ± 1351.18 & 0.25 & 3820 ± 1116.06 & 0.46 \\
    CRAZYCLIMBER & 11 & 100385 ± 23988.53 & 3.58 & 104930 ± 23118.83 & 3.76 \\
    DEFENDER & 6 & 38731 ± 13137.38 & 2.27 & 44910 ± 19215.64 & 2.66 \\
    DEMONATTACK & 17 & 817.2 ± 1218.2 & 0.37 & 1031 ± 268.88 & 0.48 \\
    DOUBLEDUNK & 28 & 15.88 ± 8.9 & 0.99 & 17.6 ± 8.49 & 1.03 \\
    ENDURO & 49 & 109.9 ± 27.42 & 0.13 & 104.4 ± 23.25 & 0.12 \\
    FISHINGDERBY & 50 & -28.9 ± 23.1 & 0.48 & -30 ± 14.18 & 0.47 \\
    FREEWAY & 18 & 27.37 ± 1.85 & 0.92 & 30.8 ± 1.47 & 1.04 \\
    FROSTBITE & 9 & 2818.1 ± 1637.63 & 0.64 & 3785 ± 861.11 & 0.87 \\
    GOPHER & 16 & 5996.6 ± 2928.52 & 2.66 & 6302 ± 3369.31 & 2.8 \\
    GRAVITAR & 47 & 1340.5 ± 890.97 & 0.37 & 1340 ± 912.63 & 0.37 \\
    H.E.R.O. & 29 & 12735.4 ± 3712.12 & 0.39 & 14048.5 ± 2446.9 & 0.44 \\
    ICEHOCKEY & 5 & 7.21 ± 5.23 & 1.52 & 12.5 ± 5.78 & 1.96 \\
    JAMESBOND & 55 & 370.5 ± 242.08 & 1.25 & 360 ± 106.77 & 1.21 \\
    KANGAROO & 35 & 441 ± 353.01 & 0.13 & 500 ± 223.61 & 0.15 \\
    KRULL & 8 & 10618.1 ± 1301.69 & 8.45 & 10957 ± 1389.92 & 8.77 \\
    KUNG-FUMASTER & 43 & 255 ± 209.46 & 0 & 270 ± 195.19 & 0 \\
    MONTEZUMA'SREVENGE & 44 & 0 ± 0 & 0 & 0 ± 0 & 0 \\
    MS.PACMAN & 12 & 1538.8 ± 391.43 & 0.19 & 2609 ± 525.12 & 0.35 \\
    NAMETHISGAME & 7 & 7679 ± 2462 & 0.94 & 9665 ± 2009.26 & 1.28 \\
    PHOENIX & 20 & 1963.9 ± 1487.39 & 0.19 & 2548 ± 2296.47 & 0.28 \\
    PITFALL & 40 & -25.24 ± 199.07 & 0.03 & -3.5 ± 8.96 & 0.03 \\
    PONG & 15 & 13.32 ± 13.31 & 0.96 & 18.4 ± 7.8 & 1.11 \\
    PRIVATEEYE & 46 & 42 ± 49.36 & 0 & 40 ± 48.99 & 0 \\
    Q*BERT & 21 & 2190.75 ± 2789.24 & 0.15 & 3387.5 ± 3334.39 & 0.24 \\
    RIVERRAID & 31 & 3780.1 ± 1494.42 & 0.15 & 4311 ± 2066.83 & 0.19 \\
    ROADRUNNER & 26 & 5409 ± 4063.84 & 0.69 & 5930 ± 6188.06 & 0.76 \\
    ROBOTANK & 19 & 9.8 ± 4.29 & 0.78 & 10.7 ± 4.54 & 0.88 \\
    SEAQUEST & 39 & 856.2 ± 345.1 & 0.02 & 1002 ± 302.85 & 0.02 \\
    SKIING & 25 & -17007.41 ± 6176.5 & 0.01 & -16110.9 ± 6255.6 & 0.08 \\
    SOLARIS & 54 & 1300.4 ± 434.15 & 0.01 & 1024 ± 360.98 & -0.02 \\
    SPACEINVADERS & 32 & 352.95 ± 198.67 & 0.13 & 396 ± 141.93 & 0.16 \\
    STARGUNNER & 30 & 4698 ± 3102.48 & 0.42 & 5100 ± 4481.52 & 0.46 \\
    SURROUND & 53 & 3.57 ± 4.2 & 0.82 & 3.2 ± 3.82 & 0.8 \\
    TENNIS & 56 & -11.9 ± 4.57 & 0.37 & -13.3 ± 5.06 & 0.33 \\
    TIMEPILOT & 2 & 11830 ± 4266.79 & 4.97 & 12710 ± 4073.68 & 5.5 \\
    TUTANKHAM & 14 & 110.42 ± 65.31 & 0.63 & 133.1 ± 59.05 & 0.78 \\
    UPANDDOWN & 13 & 19616.9 ± 10519.15 & 1.71 & 21370 ± 6814.47 & 1.87 \\
    VENTURE & 45 & 0 ± 0 & 0 & 0 ± 0 & 0 \\
    VIDEOPINBALL & 33 & 12522.37 ± 9866.3 & 0.71 & 13017.3 ± 4473.44 & 0.74 \\
    WIZARDOFWOR & 27 & 2228 ± 2246.38 & 0.4 & 2470 ± 2106.68 & 0.45 \\
    YARSREVENGE & 36 & 11192.57 ± 6595.92 & 0.16 & 12064.7 ± 8045.11 & 0.17 \\
    ZAXXON & 41 & 7033 ± 3261.04 & 0.77 & 7060 ± 2196 & 0.77 \\
    \midrule
        Average &  - & - & 0.97  & - & 1.07 \\
    \bottomrule
    \end{tabular}
    \end{adjustbox}
    \caption{Performance evaluations on 57 Atari games.}
    \label{tab:exp-jat-extend}
\end{table*}

\end{document}